\documentclass[]{fairmeta}
\usepackage{xspace}
\usepackage[dvipsnames,table]{xcolor}
\usepackage{booktabs}
\usepackage{makecell}
\usepackage{subcaption}
\usepackage{amsmath}
\usepackage{array}
\usepackage{supertabular,ifthen,multicol,caption}
\usepackage{adjustbox}
\usepackage{tipa}

\newcommand{\OmniASR}{{Omnilingual ASR}\xspace}

\newcommand{\vendorcorpus}{\OmniASR Corpus\xspace}
\newcommand{\vendoromsf}{Omnilingual + OMSF ASR\xspace}
\newcommand{\allasr}{\textsc{AllASR}\xspace}
\newcommand{\ssldata}{\textsc{SSLCorpus}\xspace}
\title{\OmniASR: Open-Source Multilingual Speech Recognition for 1600+ Languages}

\author[]{Omnilingual ASR team}
\author[\dagger]{Gil Keren}
\author[\dagger]{Artyom Kozhevnikov}
\author[\dagger]{Yen Meng}
\author[\dagger]{Christophe Ropers}
\author[\dagger]{Matthew Setzler}
\author[\dagger, 1]{Skyler Wang}
\author[2]{Ife Adebara}
\author[*]{Michael Auli}
\author[]{Can Balioglu}
\author[]{Kevin Chan}
\author[]{Chierh Cheng}
\author[]{Joe Chuang}
\author[]{Caley Droof}
\author[*]{Mark Duppenthaler}
\author[]{Paul-Ambroise Duquenne}
\author[]{Alexander Erben}
\author[]{Cynthia Gao}
\author[]{Gabriel Mejia Gonzalez}
\author[]{Kehan Lyu}
\author[]{Sagar Miglani}
\author[*]{Vineel Pratap}
\author[*]{Kaushik Ram Sadagopan}
\author[]{Safiyyah Saleem}
\author[]{Arina Turkatenko}
\author[]{Albert Ventayol-Boada}
\author[*, 3]{Zheng-Xin Yong}
\author[\ddagger]{Yu-An Chung}
\author[\ddagger]{Jean Maillard}
\author[\ddagger]{Rashel Moritz}
\author[\ddagger]{Alexandre Mourachko}
\author[\ddagger]{Mary Williamson}
\author[\ddagger]{Shireen Yates}

\affiliation[]{FAIR at Meta}
\affiliation[1]{Department of Sociology, McGill University}
\affiliation[2]{Department of Modern Languages and Cultural Studies, University of Alberta}
\affiliation[3]{Department of Computer Science, Brown University}

\affiliation[*]{Work conducted while at FAIR at Meta}

\contribution[\dagger]{Core contributors, alphabetical order}
\contribution[\ddagger]{Technical leadership and project management, alphabetical order}

\abstract{While automatic speech recognition (ASR) systems have made remarkable progress in many high-resource languages, most of the world’s 7,000+ languages remain unsupported, with thousands of long-tail languages effectively left behind. Expanding ASR coverage has long been regarded as prohibitively expensive and of limited benchmark value, further hampered by architectures that restrict language coverage to a fixed set that make extension inaccessible to most communities—all while entangled with ethical concerns when pursued without community collaboration. To transcend these limitations, this article introduces \OmniASR, the first large-scale ASR system designed for extensibility. More specifically, \OmniASR enables communities to introduce unserved languages with only a handful of their own data samples. On the modeling side, \OmniASR scales self-supervised pre-training to 7B parameters to learn robust speech representations and introduces an encoder–decoder architecture designed for zero-shot generalization, leveraging a large language model-inspired decoder to effectively exploit these representations. This capability is grounded in a massive and diverse training corpus; by combining breadth of coverage with linguistic variety, the model learns representations robust enough to adapt to previously unseen languages. Incorporating public resources with community-sourced recordings gathered through compensated local partnerships, \OmniASR expands coverage to more than 1,600 languages, the largest such effort to date—including over 500 never before served by any ASR system. Automatic evaluations show substantial gains over prior systems, especially in extreme low-resource conditions, and strong generalization to languages never encountered during training. Crucially, \OmniASR is released as a family of models ranging from compact 300M variants for low-power devices to large 7B models for maximum accuracy. Throughout the paper, we reflect on the ethical considerations shaping this design and conclude by discussing its broader societal impact. In particular, we highlight how open-sourcing models and tools can lower barriers for researchers and communities alike, inviting new forms of participation without requiring onerous expertise or heavy compute. All open-source artifacts from this effort are available at \url{https://github.com/facebookresearch/omnilingual-asr}.
}

\date{November 10, 2025}
\correspondence{Yu-An Chung at \email{andyyuan@meta.com}, Jean Maillard at \email{jeanm@meta.com}}

\metadata[Code]{\url{https://github.com/facebookresearch/omnilingual-asr}}
\metadata[Blogpost]{\url{https://ai.meta.com/blog/omnilingual-asr-advancing-automatic-speech-recognition }}

\begin{document}

\maketitle 

\newpage
\tableofcontents
\newpage

\section{Introduction}
\label{section:intro}

Automatic speech recognition (ASR) has made extraordinary strides in recent years, with state-of-the-art systems approaching human-level accuracy in many high-resource languages \citep{radford2023robust,pratap2024scaling,zhang2023google}. Yet beyond this small set lies the long tail of linguistic diversity—thousands of languages, most with little to no ASR support \citep{bartelds2023making}. Extending speech technology to this long tail is widely acknowledged as valuable, but in practice, it is rarely pursued at scale \citep{yadav2022survey}.

Researchers often shy away from long-tail ASR for a mix of practical and ethical reasons. From a practical standpoint, expanding coverage to low-resource languages can be expensive, requiring substantial engineering and data collection infrastructure for comparatively small amounts of training data \citep{hussen2025state}. Moreover, the returns are often seen as modest: a large investment may yield little improvement in benchmark performance, and the work may be perceived as less “impactful” than progress in dominant languages or novel model architectures. From an ethical standpoint, there is a concern that building technology for under-resourced communities without careful calibration risks disempowering those very communities, raising questions about language ownership and sovereignty \citep{choi2025fairness,reitmaier2022opportunities}.

While these concerns are real and deserve sustained attention, the prevailing hesitancy has important drawbacks. First, the assumption that long-tail ASR impact is minimal ignores the fact that for many communities, even modest ASR capabilities can be transformative—making oral archives searchable, enabling voice-driven interfaces in one’s own language, and contributing to the revitalization of endangered languages \citep{mainzinger-levow-2024-fine}. Second, the notion that such work lacks scientific value overlooks the unique technical challenges of the long tail: extreme data scarcity, orthographic variability, and phonetic diversity that can push the limits of model design and learning architectures \citep{imam2025automatic}. Finally, the fear of ethical missteps should be addressed not by withdrawal, but by building frameworks for social-centered and community-driven collaboration \citep{cooper2024s,reitmaier2022opportunities,wang2024human}—supported by transparent open-sourcing of models and evaluation tools to enable local adaptation and control \citep{NLLB2024,SEAMLESS2025}. Just as importantly, new architectures and design choices can be developed with community agency in mind, shifting innovation away from one-size-fits-all models toward systems that are extensible and co-shaped with the speakers who use them.

With that in mind, this paper introduces \textbf{\OmniASR}, a state-of-the-art multilingual speech recognition system that redefines how language coverage in this domain is approached. Beyond expanding to over 1,600 languages, the largest such effort to date and including more than 500 that have never been supported by any ASR system (see \Cref{sec:appending} for the full list), \OmniASR also shifts the paradigm for how \textit{new} languages can be brought into the fold. In most existing systems, languages not included at release can only be added through expert-driven fine-tuning—a path inaccessible to most communities. \OmniASR instead introduces the first large-scale ASR framework capable of extending to entirely new languages with just a few in-context examples. This capability is enabled by an encoder-decoder architecture designed for zero-shot generalization, scaling self-supervised pre-training to 7B parameters to extract speech representations, then exploiting them with a large language model (LLM)-inspired decoder. In practice, this means that a speaker of an unsupported language can provide only a handful of paired audio–text samples and obtain reasonable transcription quality—without training data at scale, out-of-reach expertise, or access to high-end compute. While zero-shot performance cannot yet match that of fully trained systems, it offers a far more scalable path to bringing new languages into digital reach.

\OmniASR also advances the state of multilingual ASR along more familiar dimensions. Its training corpus is one of the largest ever assembled for ASR in both volume and linguistic diversity, integrating publicly available datasets with community-sourced speech recordings collected through commissioned partnerships. To reach languages with little or no digital presence, we worked with local organizations who recruited and compensated native speakers, often in remote or under-documented regions. Evaluations across diverse benchmarks show consistent quality improvements over prior systems, particularly in low-resource settings, and demonstrate strong generalization to languages never encountered during training. To promote adoption in both research and deployment contexts, \OmniASR is released not as a single model but as a family—ranging from large 7B-parameter variants to compact 300M-parameter versions that can run on low-power devices “in the wild.”

By enabling the ability to support languages beyond the predefined set, at the initiative of speakers themselves, \OmniASR changes the terms of long-tail ASR. No model can ever anticipate and include all of the world’s languages in advance, but \OmniASR makes it possible for communities to extend recognition with their own data—without large-scale training or specialized expertise. This reframes ASR coverage not as a static inventory but as an extensible framework, opening space for community-driven adaptation and agency. Throughout the paper, we reflect on the ethical considerations guiding this approach, and we conclude by discussing the broader societal impact of enabling speech technology for the world’s long-tail languages.

To spur further research and enable community-driven expansion, we open-source the following at  \url{https://github.com/facebookresearch/omnilingual-asr}.:
\begin{itemize}
    \item a suite of self-supervised~(SSL) pre-trained speech models that come in 300M, 1B, 3B, and 7B parameters, all of which cover 1600+ languages suitable for fine-tuning for a wide range of downstream speech tasks and varying computational conditions;
    \item a suite of supervised connectionist temporal classification~(CTC) based ASR models fine-tuned from the SSL checkpoints suitable for basic ASR applications with strong performance;
    \item a suite of supervised LLM-based ASR models for state-of-the-art ASR performance;
    \item a zero-shot LLM-based ASR model that transcribes utterances of unseen languages using only a few examples provided at inference time;
    \item a massively multilingual ASR dataset covering over 300 languages, with an average of 10 hours of transcribed speech per language; for many languages, this represents the first ASR corpus ever built.
\end{itemize}

\section{Speech Recognition for Long-Tail Languages}
\label{section:problem}

\subsection{A Brief Overview of ASR}

ASR has long been imagined as a cornerstone of human–computer interaction, with early systems in the mid-20th century only able to recognize digits or a few carefully scripted words \citep{davis1952automatic}. Over the decades, research steadily expanded the scope of what ASR could do, from isolated command-and-control vocabularies to continuous recognition of natural speech \citep{young1996review}. A critical driver of this progress was the availability of benchmark datasets that allowed researchers to measure advances and refine algorithms in widely spoken languages like English \citep{garofolo1993timit}. By the 2010s, with the rise of deep learning, ASR reached a turning point: feedforward deep neural networks (DNNs) and later recurrent neural networks (RNNs) drastically improved acoustic modeling, while sequence-to-sequence and attention-based architectures laid the foundation for fully end-to-end ASR systems \citep{chorowski2015attention,graves2014towards}. Large public corpora like LibriSpeech \citep{panayotov2015librispeech}, derived from audiobooks, further accelerated progress by standardizing evaluation in English.
Systems trained on large amounts of labeled data began approaching human-level accuracy for certain high-resource languages, and speech technology entered everyday applications from voice assistants to automated captioning \citep{radford2023robust}.

The more recent wave of progress has been propelled by scaling—both in terms of training data and model architectures. Datasets such as MLS~\citep{pratap2020mls}, VoxPopuli~\citep{wang2021voxpopuli}, MSR \citep{li2024msr} and Granary \citep{koluguri2025granary} have substantially increased the amount of transcribed speech available for training, though these advances have been directed mostly at languages which were already high-resource. Efforts to include lower-resource languages have accelerated in recent years, with datasets such as BLOOM \citep{leong2022bloom} covering 56 languages, Speech Wikimedia \citep{gomez2023speech} reaching 77, and YODAS \citep{li2023yodas} spanning 140. Yet despite these expansions, the distribution of data remains heavily skewed, and only a handful of recordings exist for many of the most under-served languages. A broader coverage of nearly 700 languages is offered by CMU wilderness \citep{black2019cmu}, which was derived from publicly available Bible recordings and therefore lacks diversity in domain, reading style, and speakers. An analogous effort that is primarily restricted to the religious domain is the MMS dataset \citep{pratap2024scaling}, reproduced in its untranscribed part by \citet{chen2024robustspeechrepresentationlearning}, representing the largest coverage to date with over 4,000 languages. Of particular note are projects such as VAANI \citep{vaani2025}, which is dedicated to the collection of natural speech in over 100 languages from the Indian subcontinent, and African Next Voices \citep{anv-za,anv-ke,anv-et-amh,anv-et-sid,anv-et-wal,anv-et-tig,anv-et-orm}, which focuses on providing large, high-quality and culturally rich datasets for African languages. Common Voice \citep{ardila2020common}—maintained by the Mozilla Foundation and curated by a large network of volunteers—currently spans approximately 130 languages and stands out as the most extensive and widely utilized datasets.

Advancements made in self-supervised learning have further reshaped the field. More specifically, models like wav2vec~2.0~\citep{baevski2020wav2vec} demonstrate how massive amounts of unlabeled audio could be leveraged to learn powerful speech representations, drastically reducing the need for labeled data. This paradigm enabled breakthroughs such as the Universal Speech Model by \citet{zhang2023google}, pre-trained on 12 million hours of unlabelled speech spanning over 300 languages and fine-tuned on a smaller labeled dataset, and the MMS model of \citet{pratap2024scaling}, which extended coverage beyond 1,100 languages through large-scale pre-training. Self-supervision can also improve the language modeling or text generation component of ASR systems, including in multilingual settings, as demonstrated by works by \citet{babu2021xls}, \citet{bapna2022mslam}, and \citet{pratap2024scaling}. Moreover, architectural innovations can allow models to transcribe languages unseen during training. For instance, \citet{li2022asr2k} propose an approach based on mapping the output of an 8-language multilingual model to language-specific phonemes, a method extensible to any unseen languages which have n-gram statistics, though limited by the reliability of phoneme mappings for low-resource languages. Building on this, \citet{zhao2025scaling} remove the intermediate phone representations and instead adopt a romanization-based encoding, achieving lower error rates.  Although recent advances in language adaptation and zero-shot capabilities of large language models show promise \citep{yong2023bloom}, these gains have so far accrued mainly to high-resource languages \citep{ahuja2023mega,bang2023multitask,asai2024buffet,ochieng2025beyond}.

\subsection{Overcoming challenges to Long-Tail ASR}

From above, we see that despite recent achievements in the field of ASR, the benefits remain concentrated in a relatively small subset of high-resource languages, leaving the vast majority of the world’s linguistic diversity unsupported. Understanding why such an important problem is rarely undertaken at scale requires unpacking the practical, scientific, architectural, and political barriers that have kept many long-tail languages on the margins of ASR development. Below, we outline some of these hurdles.

\textbf{Practical barriers.} Collecting training data for low-resource languages is resource-intensive. Unlike high-resource languages, which have vast amounts of texts and transcribed speech available, many long-tail languages require costly, ground-up data creation \citep{abraham2020crowdsourcing,besacier2014automatic}. This often involves recruiting native speakers, designing orthographic conventions, and collecting high-quality audio in settings where infrastructure may be limited. The effort is large, yet the resulting datasets are comparatively small, making them less attractive for institutions prioritizing efficiency or scale \citep{blasi2021systematic}.

\textbf{Scientific disincentives.} In the research community, progress is typically measured by benchmarks and leaderboard gains. Improving ASR for a long-tail language rarely moves the needle on widely used benchmarks, and therefore can be perceived as less “impactful” or publishable \citep{mainzinger-levow-2024-fine}. The challenges are also technically demanding: extreme data scarcity, phonetic diversity, and variable orthographies stretch existing architectures beyond their tested limits \citep{adda2016breaking,joshi2020state}. These are precisely the kinds of challenges that could advance the science of ASR, but in practice they often push researchers toward safer ground.

\textbf{Architectural limitations.} Existing ASR systems generally treat language coverage as fixed at release. If a language is not included in training, extending support typically requires expert-driven fine-tuning with large compute resources and specialized expertise—an approach inaccessible to most communities \citep{imam2025automatic}. This lack of extensibility effectively prevents many groups from bringing their languages into digital spaces, slowing progress toward inclusive ASR.

\textbf{Ethical and political complexities.} Long-tail languages are deeply entangled with questions of identity, ownership, and sovereignty. Building ASR systems without community involvement risks creating extractive dynamics \citep{bird2024must}, where outside institutions “take” language data without returning meaningful benefits to speakers. Concerns about appropriation or misuse have led some researchers to avoid long-tail ASR altogether, fearing that well-intentioned efforts might inadvertently disempower the very communities they aim to support \citep{choi2025fairness,cooper2024s}.

While these practical and ethical concerns explain the historical neglect of long-tail languages, leaving them unsupported is far from a neutral choice. The lack of ASR capacity has tangible consequences for the communities situated at the margins \citep{joshi2020state}. Many of these languages are primarily oral, with few standardized orthographies or written resources. Without ASR, oral archives—from folktales to political speeches—remain locked in raw audio, inaccessible to researchers, educators, or even community members seeking to preserve and circulate their own heritage. In more everyday terms, the absence of speech technology excludes entire populations from tools that dominant-language speakers take for granted: dictation, search, subtitling, or voice-based accessibility services \citep{mainzinger-levow-2024-fine}. This exclusion is not simply technical; it reinforces digital hierarchies in which only speakers of globally dominant languages can fully participate in an increasingly voice-driven digital ecosystem \citep{SEAMLESS2025}. For minority communities, the effects can be even more acute, as the lack of technological affordances accelerates language shift: younger speakers may turn toward dominant languages that provide digital tools, leaving their heritage languages further marginalized \citep{kornai2013digital}.

This current effort hopes to transcend these barriers by recognizing that inaction perpetuates inequality. Not building ASR for long-tail languages is itself a decision—one that deepens digital divides and risks silencing already vulnerable voices. To counter this, our approach prioritizes community partnerships, ensuring that the extension of ASR coverage is developed collaboratively with local actors. By working directly with communities, compensating native speakers for speech data, and enabling local adaptation through open-source release, \OmniASR aims not only to expand technical coverage but to lay the groundwork for more inclusive, community-driven participation in the speech technology ecosystem.

\section{Data and Language Coverage}
\label{section:data}

Building a system that can recognize and transcribe speech across more than 1,600 languages first required the largest and most diverse ASR training corpus assembled to date. Achieving this breadth meant integrating resources from multiple domains: existing public datasets, internal collections developed for prior multilingual ASR systems, and crucially, community-sourced speech recordings that extend coverage into languages with little or no prior digital footprint. In this section, we provide additional information about language coverage and break down the training corpus creation process.

\subsection{Referring to Languages}
In the absence of a strict scientific definition of what constitutes a \textit{language}, we adopted a practical convention: treating as candidate languages those linguistic entities—\textit{languoids}, following \citet{good2006modeling}—that have been assigned their own ISO 639-3 codes.

We acknowledge that language classification in general, and the attribution of ISO 639-3 codes in particular, is a complex process, subject to limitations and disagreements, and not always aligned with how native speakers themselves conceptualize their languages. To allow for greater granularity when warranted, ISO 639-3 codes were complemented with Glottolog languoid codes \citep{hammerstrom2024glottolog}. For example, we preserved the distinction between the Vallader and Sutsilvan varieties of Romansh, following the practice of the Mozilla Common Voice community, by using the Glottocodes \texttt{lowe1386} and \texttt{suts1235}. In the rare cases where Glottolog’s classification is known but actively disputed by the speaker communities we worked with, we supplemented ISO 639-3 codes with community-supported languoid names; for instance, by adopting the IANA language variant subtags \texttt{gherd} and \texttt{valbadia} for Ladin.

Due to the written component of the ASR task, careful attention was also paid to languages with multiple writing systems. Accordingly, all languages supported by our model are associated with one or more ISO 15924 script codes. Take Mandarin, for example, we use \texttt{cmn\_Hant} to denote Mandarin Chinese in the traditional script and \texttt{cmn\_Hans} for the same language in the simplified script. Where additional variants are needed, we extend this system; for example, \texttt{roh\_Latn\_suts1235} identifies the Sutsilvan Romansh languoid written in the Latin script.

\subsection{Defining Language Coverage}\label{sec:data:coverage}
For ASR applications, at least some of the training data must consist of speech recordings paired with transcripts. The first steps in defining language coverage are therefore to ensure, first, that the language candidates are spoken, and second, that they have an established writing system. Both points warrant brief elaboration.

First, the ISO 639-3 inventory (with more than 7,000 codes) includes roughly 150 signed languages. Because these are not spoken, they cannot be directly included in ASR applications. Second, the availability and classification of writing systems is far from straightforward. It is not a simple dichotomy between written and unwritten languages. Some languages consistently employ a single writing system, while others have used multiple systems either historically or concurrently. In certain cases, these practices are well documented; in others, information is incomplete or missing. For instance, ScriptSource\footnote{\url{https://scriptsource.org/entry/wekytddkkc} (retrieved 2025-08-19)} reports 2,586 languages with insufficient information on their scripts. This does not imply that the languages in question are unwritten, but it does highlight the challenges of securing textual data for them.

Our approach was to include only languages with at least one established writing system. By “established,” we mean a form of writing that is in frequent use, intelligible to the speaker community, and ideally described in formal resources such as dictionaries or grammars. This excludes transcriptions in the International Phonetic Alphabet\footnote{International Phonetic Alphabet} or idiosyncratic note-taking systems, which do not constitute stable or widely recognized orthographies.

Beyond the above considerations, additional steps were taken to define the scope of our language coverage while avoiding overlapping or redundant inclusion. Overlap can occur through macrolanguage codes or through duplication with already available data. Macrolanguage codes are a known feature of ISO 639-3. The standard defines 63 such codes, which may be used either to group related varieties or as a placeholder where more specific identification is unavailable. However, many macrolanguage codes are overly broad and often redundant. For example, the macrolanguage code \texttt{msa} for the Malay group of languages encompasses 36 other ISO 639-3 codes, including Indonesian and Minangkabau. To minimize ambiguity, such macrolanguage codes were excluded wherever possible. Lastly, we also deprioritized languages already covered in prior ASR work, such as \citet{pratap2024scaling}, on which \OmniASR builds. Finally, constructed languages and languages classified by UNESCO as extinct were also deprioritized, as neither provide a viable basis for ASR applications.

\subsection{Dataset creation}

Building \OmniASR involved compiling the largest linguistically diverse speech dataset ever created. In this section we detail the extensive efforts undertaken to assemble existing resources and develop new ones through partnerships and commissioning.

\subsubsection{Existing ASR Data}
\label{sec:existing_asr_sources}

We assembled training data from a large number of existing open-source datasets: ALFFA \citep{abate2005alffa,gelas2012alffa,gauthier2016alffa}, LibriSpeech ASR \citep{panayotov2015librispeech}, the South African language data of \citet{vanniekerk2017rapid}, ASR and TTS data by \citet{kjartansson2018crowd}, \citet{kjartansson2018tts} and \citet{he2020open}, CSS10 \citep{park2019css10}, FOSD \citep{fosd}, Zeroth Korean dataset,\footnote{\url{https://github.com/goodatlas/zeroth}} Burmese Speech Corpus \citep{oo2020burmese}, Common Voice v22 \citep{ardila2020common}, VoxPopuli \citep{wang2021voxpopuli}, VoxLingua-107 \citep{valk2021slt}, RuLS,\footnote{\url{https://www.openslr.org/96/}} the Kokoro Speech Dataset,\footnote{\url{https://github.com/kaiidams/Kokoro-Speech-Dataset}} MLS \citep{pratap2020mls}, {S}amr{\'o}mur \citep{mollberg2020samromur}, the Kazakh Speech Corpus \citep{khassanov2021crowdsourced}, iMaSC \citep{gopinath2022imascic}, ParlaSpeech-HR \citep{ljubesic2022parlaspeech}, NPSC \citep{solberg2022norwegian}, FLEURS \citep{conneau2023fleurs} and NaijaVoices \citep{emezue2025naijavoices}.

We supplemented these sources with additional ASR data, coming from an internal dataset of publicly available speech recordings paired with transcriptions, and a number of commercially-available licensed datasets including the 17 language packs from the IARPA Babel program \citep{gales2014babel}.

Finally, we integrated these resources with datasets shared from partners taking part in our Language Technology Partner Program, an effort intended to offer opportunities for interested members of the public to contribute to AI language technologies, with a particular focus on under-served languages. Participating members were able to access technical workshops led by our research team, learning how to leverage open-source models to build language technologies for their languages.

\subsubsection{Partner-Created ASR Data}

To support the development of speaker-centric ASR datasets, we provided funding and additional resources for several collaborative initiatives that placed native speakers and local communities at the center of the process, ensuring that the data collected was truly reflective of their linguistic and cultural input.

One such key effort is the African Next Voices project, a consortium led by Maseno University in Kenya, University of Pretoria in South Africa and Data Science Nigeria, aiming to bridge the technological divide in speech technologies for African languages and to promote equitable AI development across the continent.  This project—which is supported by the Gates Foundation—ultimately aims to provide tens of thousands of hours of ASR data for up to twenty of the continent's most spoken languages. The significant progress from this ongoing initiative is well documented in numerous scientific papers and open-source artifacts \citep{anv-za,anv-ke,anv-et-amh,anv-et-sid,anv-et-wal,anv-et-tig,anv-et-orm,anv-kin,anv-swh}.

Additionally, we provided support to the Open Multilingual Speech Fund by Mozilla Foundation's Common Voice \citep{ardila2020common}. This empowered over 170 new language communities to join the project. This support for community-centered open data work has enabled the number of communities participating in Common Voice to more than double. It brings the Common Voice corpus to well over 300 languages, helping to enrich linguistic diversity in technology for everyone.

Finally, we supported the Lanfrica/Naijavoices initiative,\footnote{\url{https://naijavoices.com/}} which resulted in the creation of new datasets for 11 African languages (Bainouk-Gunyaamolo, Balanta-Kentohe, Bube, Fang, Igala, Central Kanuri, Karon, Nupe-Nupe-Tako, Upper Guinea Crioulo, Serer and Urhobo) with a focus on high-quality, culturally representative, and demographically diverse content.

\subsubsection{Commissioned ASR Data: The \vendorcorpus}
\label{sec:vendor_corpus_desc}

In addition to drawing on the aforementioned resources, we commissioned a tailored set of recordings and transcriptions to strengthen the corpus. This step ensured that the model would be trained on domain-diverse, high-quality spontaneous speech spanning a broad range of languages. By proactively filling gaps left by prior datasets, we aimed to create a resource that not only meets the immediate needs of this project but also enhances the model’s long-term adaptability. As we show in \Cref{sec:zs_asr,sec:zs_icl}, this diverse foundation is already demonstrating its value by facilitating cross-lingual transfer through zero-shot generalization. Below, we document the steps taking to develop the \vendorcorpus, all of which is open-source and be made publicly available.

\begin{figure*}[t!]
    \centering
    \begin{subfigure}[t]{0.45\textwidth}
        \centering
        \includegraphics[height=1.5in]{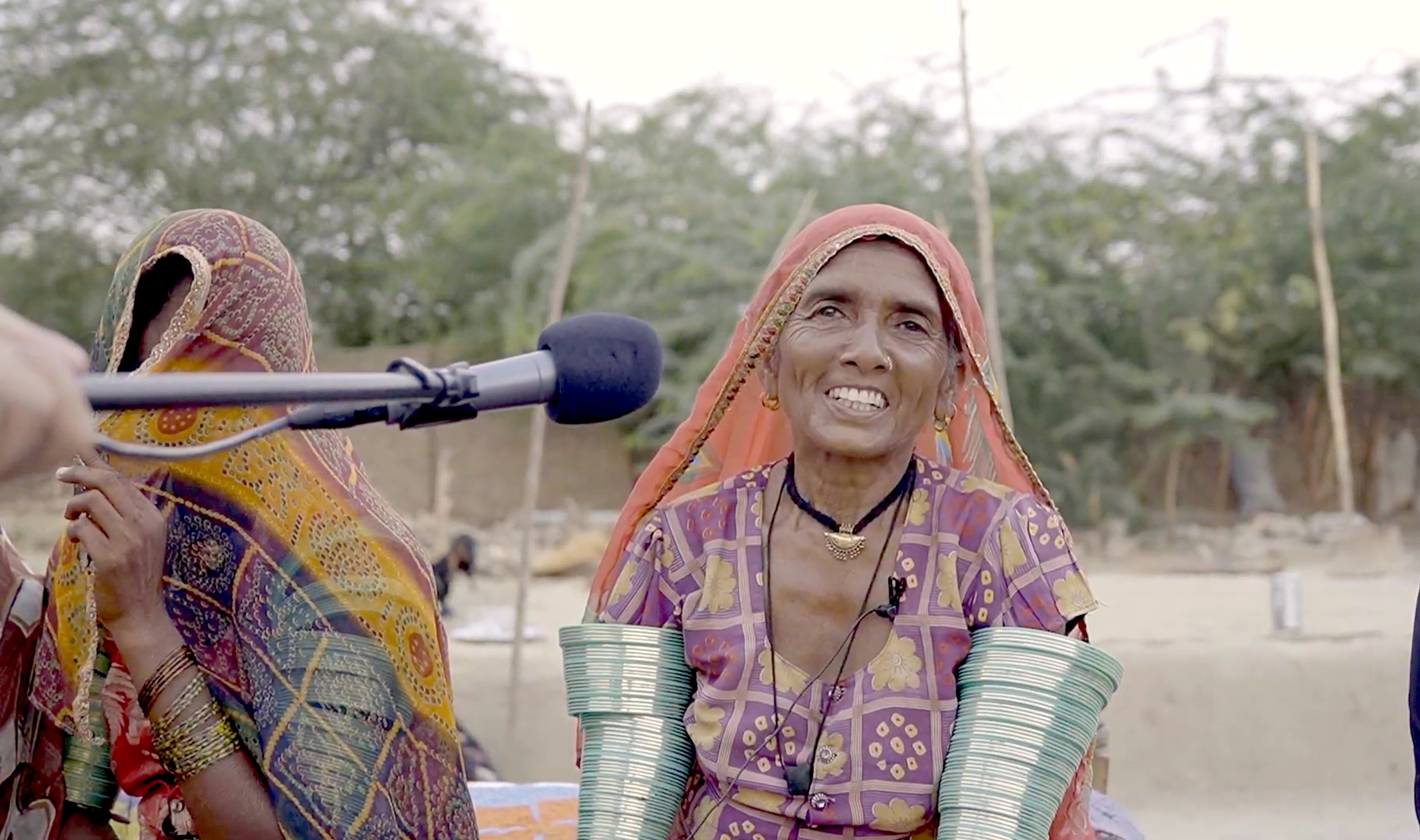}
        \caption{Local participants contributing to corpus creation efforts in Pakistan.}
    \end{subfigure}%
    ~ 
    \begin{subfigure}[t]{0.45\textwidth}
        \centering
        \includegraphics[height=1.5in]{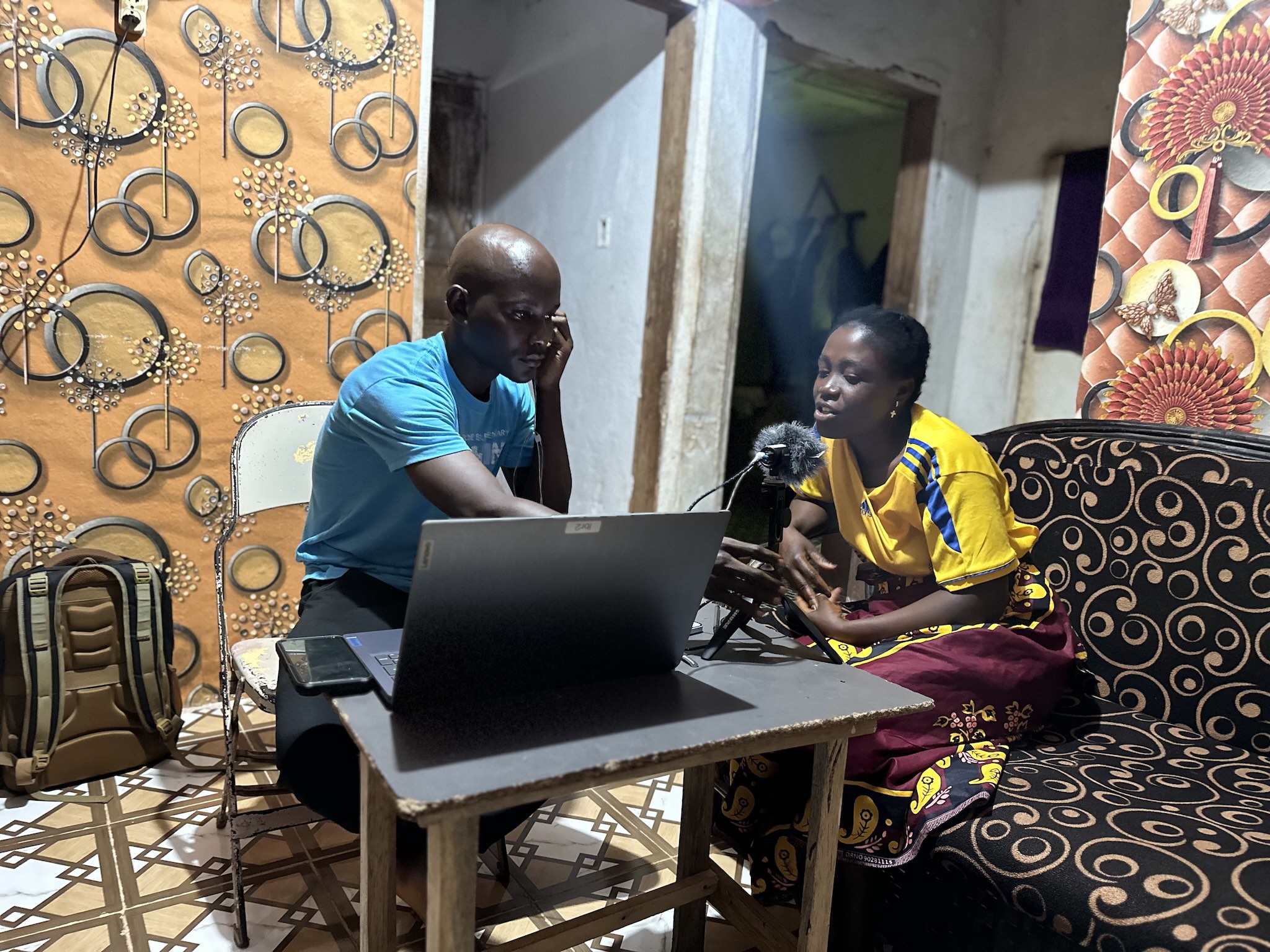}
        \caption{Local participants contributing to corpus creation efforts in Liberia.}
    \end{subfigure}

    \begin{subfigure}[t]{0.7\textwidth}
        \centering
        \includegraphics[height=1.5in]{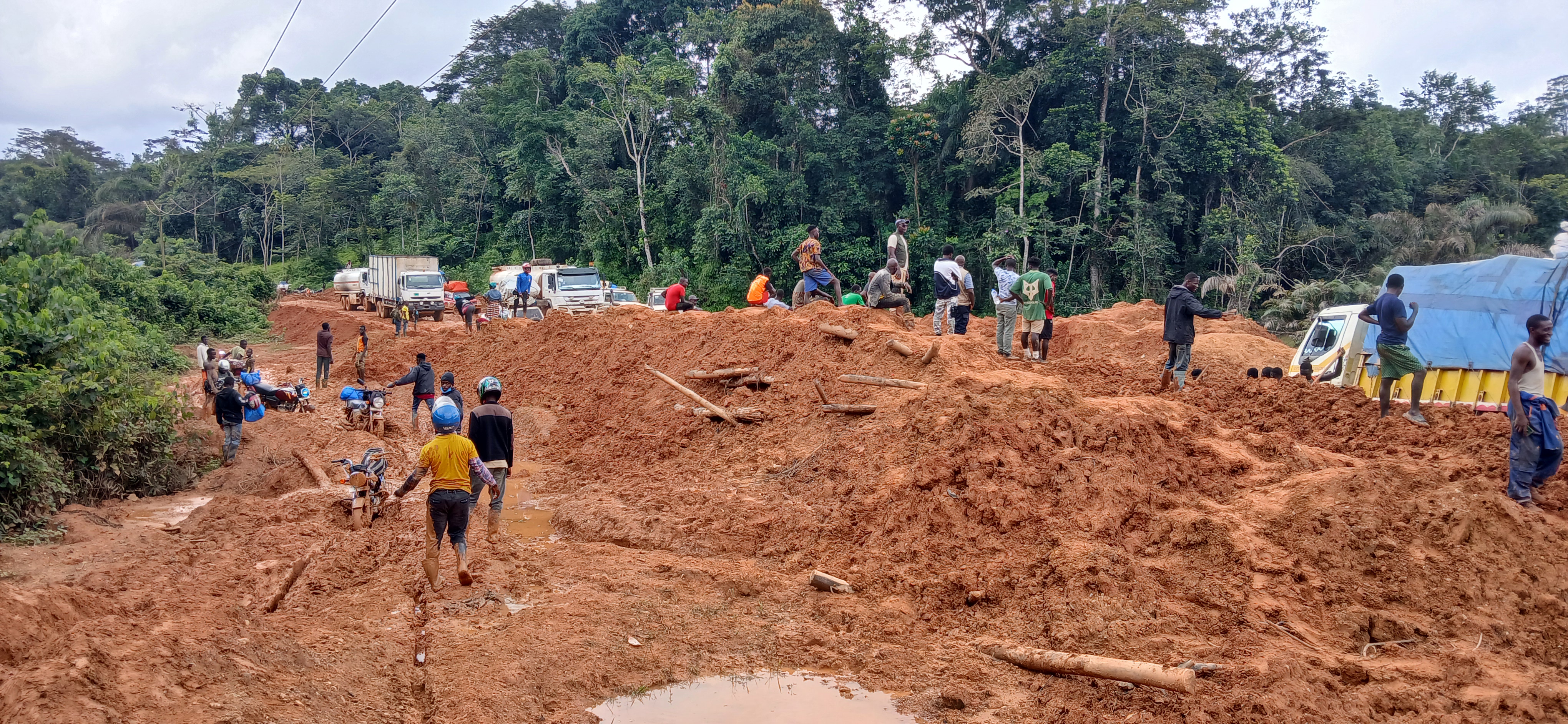}
        \caption{Example of the difficult travel conditions encountered during fieldwork.}
    \end{subfigure}

    \caption{Photographs documenting key moments from the global collection of speech data that produced the \vendorcorpus.}
    \label{fig:data_collection}
\end{figure*}

\textbf{Prompt design.} Our initial goal was to commission the collection of 10 hours of speech from 10 different native speakers (1 hour per speaker) in each for roughly 350--400 languages, paired with corresponding transcripts. To elicit naturally occurring language grounded in speakers’ experiences while avoiding personal information, we developed survey-style prompts such as \textit{Is it better to have a few close friends or many casual acquaintances? Why?} Vendors were provided with a pool of more than 1,500 such prompts, ensuring sufficient material for one hour of naturally-occurring speech. The prompt set was made available in English and six pivot languages (French, Indonesian, Italian, Mandarin Chinese, Portuguese, and Spanish).

Importantly, we deliberately over-supplied prompts—far more than any speaker would need for a single session. This decision served several purposes. First, no single set of questions can feel equally relevant worldwide; by offering breadth, we allowed participants to skip prompts they found uncomfortable or uninteresting. Second, the abundance of options let speakers guide the recordings toward topics they cared about, fostering engagement and spontaneity. In practice, many participants moved fluidly between prompts and their own digressions. For example, one speaker began with a lighthearted role-play prompt about imagining life as a bird and ended with a detailed reflection on the nesting habits of local bird species. This design ensured that our dataset was not only broad and balanced but also enriched with authentic, culturally grounded, and participant-driven speech.

\textbf{Native speaker availability.} In practice, it was not always possible to follow the initial collection plan exactly. First, suitable speakers could not be found in all languages within the specified time frame. In some cases, this meant that the 10-speaker target was not met, reducing the total amount of collected recordings and transcripts. In others, the shortfall was offset because available speakers recorded more than one hour each, allowing the 10-hour target to be met even without 10 distinct contributors. A further set of languages had speakers recruited but did not complete the full collection in time for inclusion in the training mix; nonetheless, we release those recordings and transcripts as part of the final open-source dataset. Finally, in a positive deviation from plan, vendors were able to document established writing systems for some languages not initially listed as candidates, and proceeded to collect speech recordings and transcripts for them as well. \Cref{tab:all_asr_dataset} summarizes basic statistics on all training data, including the commissioned data collection to date (\vendorcorpus).

\textbf{Recordings.} Participants were provided with prompts (or, in some cases, had prompts read aloud to them) and asked to respond. Prompts could be delivered either in participants’ native languages or in a second language in which they were proficient, but responses were to be given in their native languages, spoken naturally and at a normal pace—neither rushed nor artificially slow. When references to foreign terms were needed, participants were encouraged to pronounce them as they ordinarily would when speaking with fellow native speakers. Finally, participants were instructed to avoid sharing any personally identifiable information (PII), with a full list of items considered so provided in \Cref{appx:guideline:record}.

\textbf{Transcripts.} For the purpose of building ASR datasets, speech recordings must be paired with accurate transcriptions. We define accuracy here in two dimensions: first, transcriptions should be produced in an established writing system for each language (see \Cref{sec:data:coverage}); second, they must adequately reflect the characteristics of naturally-occurring spontaneous speech.

Unlike scripted or prepared speech, spontaneous speech exhibits disfluencies (repetitions, false starts, repairs, or incomplete sentences). These occur alongside non-verbal vocalizations such as fillers, laughter, breathing sounds, or coughs. To ensure faithful transcripts, such events must be annotated, along with occasional non-vocal sounds and background noise. For this purpose, participants were asked to use special tags—\texttt{<laugh>}, \texttt{<hesitation>}, \texttt{<unintelligible>}, and \texttt{<noise>}. Further details on tag usage are provided in the transcription guidelines (see \Cref{appx:guideline:transcript}).

In addition to typical challenges that stem from the complexity of accurate spontaneous speech transcription in any language, more specific challenges also arise when attempting to transcribe low-resource languages, many of which are facing intergenerational disruption \citep{fishman1991reversing}. It is not uncommon for native speakers of disrupted languages to reside in more rural areas, where getting access to digital devices that produce and store machine-readable transcripts can be a challenge. Even when such devices are available, they may not support the relevant script or orthography.  It might also happen that speakers who have native mastery of the spoken language do not feel as comfortable with its written form. For these reasons, transcripts were not always produced by the speakers themselves. In some cases, they were prepared by on-site typists; in others, handwritten notes were later digitized off-site. Each degree of separation from the original speaker introduced additional challenges to achieving transcription accuracy.

\begin{figure}[!htb]
    \centering
    \includegraphics[width=0.9\textwidth]{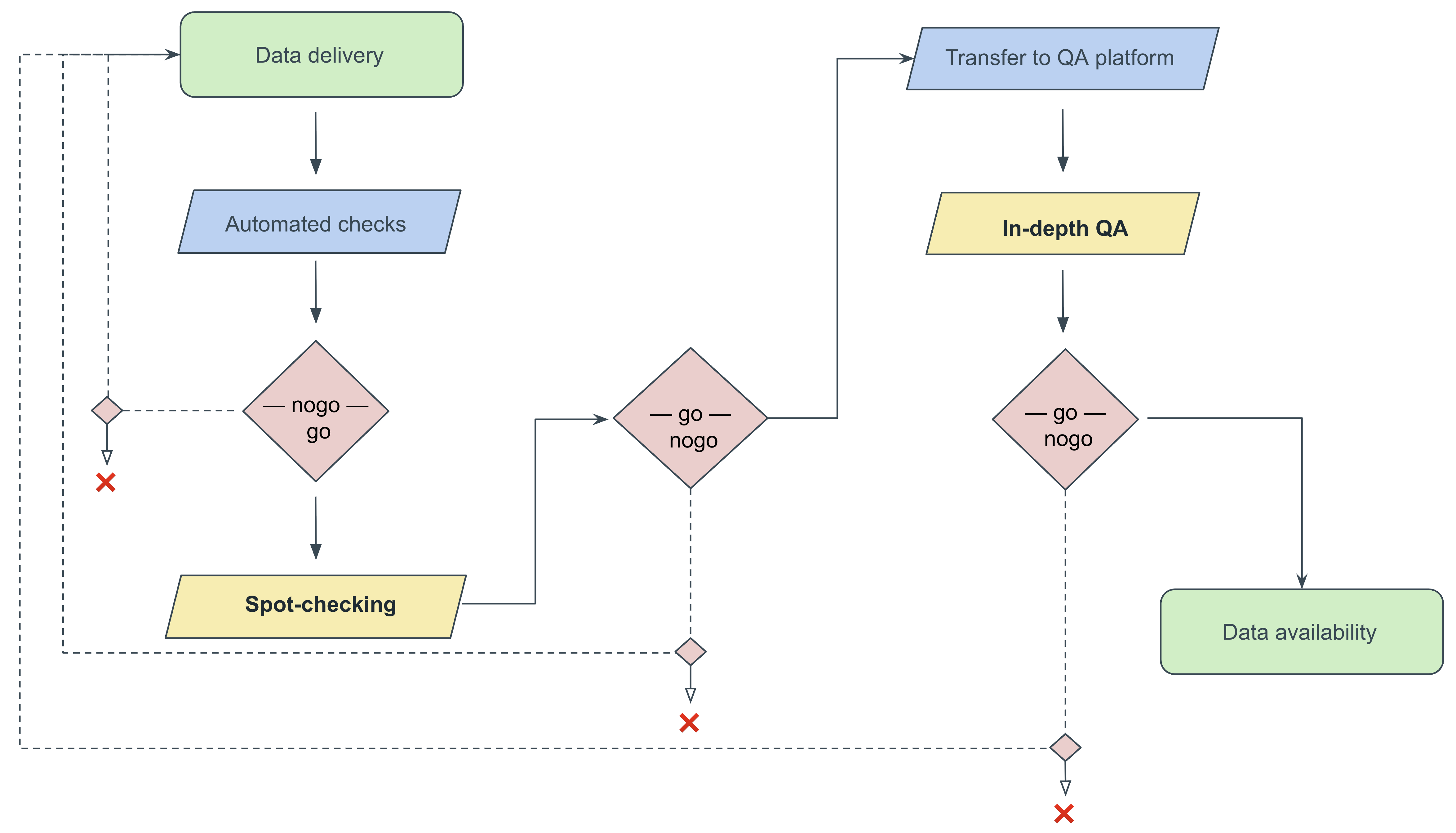}
    \caption{
        Commissioned data quality-assurance workflow.}
    \label{fig:data:qa_workflow}
\end{figure}

\textbf{Quality assurance (QA). }
\Cref{fig:data:qa_workflow} shows the process by which the quality of the commissioned data was controlled. First, at the partial delivery stage, files were automatically screened for major quality flaws, such as corruption during transfer, unexpected duration, or excessive noise levels. A small number of files per language were also manually inspected by linguists, prioritizing those files that returned unexpected automated check results. After these initial rapid quality checks, feedback was communicated to vendors for easier root-cause identification and error correction. Then, at the final delivery stage, both speech and text data were uploaded to a specifically designed QA platform, and were inspected by trained QA technicians.

The QA platform enabled technicians to access each speech recording alongside its corresponding transcript within a single interface, which also displayed the quality questionnaire they were required to complete. The primary objectives of this task were to detect potential errors and classify them as either minor or critical. \Cref{tab:qa_errors} provides definitions for the most common error types in both categories, while a detailed description of the QA procedure and error taxonomy for speech recordings and transcripts is provided in \Cref{appx:guideline:qa}.
\begin{table}[!ht]
    \centering
    \begin{tabular}{p{3cm}p{6cm}p{6cm}}
    \toprule
     \textbf{\small{Category}}  & \textbf{\small{Critical example}} & \textbf{\small{Minor example}} \\
     \midrule
     \small{Human vocal noise} & \small{Second voice in the background} & \small{N/A (This error is always critical)} \\
     & \small{Singing in the background} &  \\
    \midrule
    \small{Cutoff} & \small{Speech is cut off at either end of the recording} &  \small{N/A (This error is always critical)}\\
     \midrule
     \small{Background noise} & \small{Rooster crowing} & \small{Occasional mild coughing} \\
     & \small{Street noise, car honking} & \small{Occasional mild coughing} \\
     & \small{Bird chirping} & \small{Mild breathing sound}\\
     & \small{Strong wind} & \\
     \bottomrule
    \end{tabular}
    \caption{Description of the error categories used for in-depth quality assurance (audio files)}
    \label{tab:qa_errors}
\end{table}

Every language in the \vendorcorpus went through at least the first step of human review (small-scale inspection), and 279 languages went through in-depth inspection. When rework was possible, quality issues were mitigated. In other cases, the portion of the data that did not meet quality requirements was excluded.  

The QA process was instrumental in detecting and mitigating issues in data deliveries. Considering both minor and critical errors, the most frequent problems in audio files were long silences and background noise, while transcript files most often exhibited spelling inconsistencies and mismatches. Spelling inconsistencies are common in low-resource languages, where orthographies are not standardized in the same way as they are in high-resource languages. Mismatches between speech and transcripts, by contrast, are more serious but relatively straightforward to fix when identified early, as they usually reflect file misalignments rather than transcription errors per se.

Focusing on critical errors specifically, \Cref{tab:qa_results} provides a more detailed breakdown of the six most prevalent categories. After long pauses, the most prevalent critical issues in speech recordings were cutoffs and human vocal noises. Cutoffs are likely the result of the recording equipment being mishandled, while vocal noises typically arose from audible human voices captured in the background.

\begin{table}[!ht]
    \centering
    \begin{tabular}{p{4cm}>{\raggedleft}p{3cm}p{4cm}p{3cm}}
\toprule
\multicolumn{1}{l}{\textbf{\small{Critical audio issues}}} & \multicolumn{1}{c}{\textbf{\small{Percentage of files}}} & \multicolumn{1}{l}{\textbf{\small{Critical transcript issues}}} & \multicolumn{1}{c}{\textbf{\small{Percentage of files}}} \\
\midrule
\small{Pause / Silence} & $27.25\%$ & \small{Mismatch} & $51.18\%$ \\
\small{Cutoff} & $15.62\%$ & \small{Incomplete or summarized} & $21.97\%$ \\
\small{Human vocal noise} & $10.62\%$ & \small{Wrong writing system} & $10.51\%$ \\
\small{Background Noise} & $9.42\%$ & \small{Wrong tags} & $8.20\%$ \\
\small{Unnatural speech} & $9.05\%$ & \small{Numbers} & $1.97\%$ \\
\small{Low volume} & $5.31\%$ & \small{Inconsistent tagging} & $1.44\%$ \\
\bottomrule
    \end{tabular}
    \caption{Most prevalent critical quality issues in speech and transcripts files}
    \label{tab:qa_results}
\end{table}

\textbf{Validation.}
\citet{kreutzer2022quality} show that a common quality issue in large, massively multilingual datasets stems from dataset mislabeling; i.e., the misattribution of language codes to some subsets of the data corpus. Such misattributions can be caused by several factors: for example, the use of both a private code and an attributed ISO code for the same language. Languages are often also known by different names in English and other languages, and even by different autonyms within their own native speaker groups. When the name of a language appears to be absent from the list of language names that correspond to ISO codes, it is tempting to create a private code without realizing that the language already has its ISO code under a slightly (or not so slightly) different name. Another type of code misattribution can come from a confusion between the code for a spoken language and the code for a sign language by a similar name (e.g., Hausa [hau] and Hausa Sign Language [hsl]).

To mitigate language code misattribution issues in the commissioned data, a validation project was set up whereby a small portion of the data collected by one vendor for a particular language was analyzed by a different vendor. The volume per language ranged between 1 to 5 audio files and up to 10 transcripts. For each sample audio and transcript file, proficient speakers of the target language were asked to determine whether the sample represented acceptable spoken or written forms of their language. Vendors were given additional guidance as to potential miscommunication due to the language naming discrepancies previously mentioned, as well as to discrepancies in the use of the terms \textit{language} and \textit{dialect}. 

The language code validation process was applied to 206 
languages, and allowed us to identify instances of misattributed language codes in 20 
languages. These findings further underscore the significant challenges associated with collecting accurate data for Arabic and Fula languages in particular. The validation process also indirectly helped identify and correct a general language code attribution error for [\texttt{zga}]. For clarity, this language code validation step only constitutes additional due diligence on a very small portion of the datasets. The results of this process, whether negative or positive, should not lead to generalizations about entire datasets. Nevertheless, they provided additional insights into the quality of the commissioned data and into opportunities for improvement. 

\subsubsection{Pre-training data}
\label{sec:ssl_data}

As we will go into details in \Cref{sec:w2v2}, \OmniASR is built on a massively multilingual speech encoder capable of producing high-performing cross-lingual speech representations. 
Training this encoder required a large-scale corpus of unlabeled speech. To construct it, we combined all the sources described in the preceding sections that were available when encoder training began. This phase predated the fine-tuning of the ASR models by several months, as well as the full delivery of our \vendorcorpus  and several partner-contributed ASR datasets.  To further expand coverage, we supplemented these resources with a large-scale internal collection of unlabeled speech.  The final pre-training dataset comprised $3.84\text{M}$ hours of speech across $1{,}239$ languages, in addition to another $460\text{K}$ hours of speech for which no language identification was performed. 

\subsection{ASR Data Preparation and Cleaning}

Concretely, we first split the text using the \textbf{sat-12l-sm} SAT model from \citet{frohmann-etal-2024-segment}. By leveraging its splitting probability outputs, we ensured that text segments remained shorter than 200 characters. Annotators had often already inserted sentence boundaries, and SAT segmentation typically rediscovered this structure. However, for languages entirely out of SAT’s training domain and without sentence-level annotations, segmentation was instead driven by the maximum length constraint, without necessarily following sentence structure. Next, we applied a forced-alignment algorithm to obtain corresponding audio segments, following the procedure described in \citet{pratap2024scaling}. If some audio segments remained too long ($>50,\text{s}$), we reapplied the split-align operation with a reduced maximum text-segment length. Conversely, if audio segments were too short ($<2,\text{s}$), they were merged with the nearest neighboring segment. Several iterations of split/merge ensured that final segments fell within the target range of $[2,\text{s}, 50,\text{s}]$. Finally, we note that no utterance-level segmentation was performed on existing public datasets such as FLEURS, MLS, or Babel.

After utterance-splitting, we applied WER-based filtering on the \vendorcorpus to remove misaligned audio–text pairs. Such problematic examples were rare and typically arose either from erroneous reference transcripts or pathological edge cases in the segmentation/alignment pipeline. For curation, we used a 7B CTC model trained on a subset of available ASR data (excluding MLS, which does not contribute to lower-resource language coverage). We computed WERs for each utterance in the \vendorcorpus datasets, then conducted qualitative analyses within each datasource to establish source-specific thresholds. Our philosophy was to apply minimal filtering, retaining as much data as possible while removing only clearly erroneous pairs. ~\Cref{sec:appendix_wer_filtering} provides the thresholds used as well as examples of filtered misalignments.

Finally, we constructed a character-based tokenizer by taking the union of all characters across the entire ASR dataset. This inventory was manually cleaned to remove obvious artifacts (e.g., punctuation, emojis) and extremely rare characters (occurring fewer than five times across the corpus) in order to limit vocabulary size. The resulting tokenizer contained 9,812 symbols. We then applied it to filter out degenerate transcripts containing >=15\% unknown tokens.

\subsection{Final Datasets}
\label{sec:final_data}

\begin{figure}[ht!]
  \centering
  \includegraphics[width=0.8\textwidth]{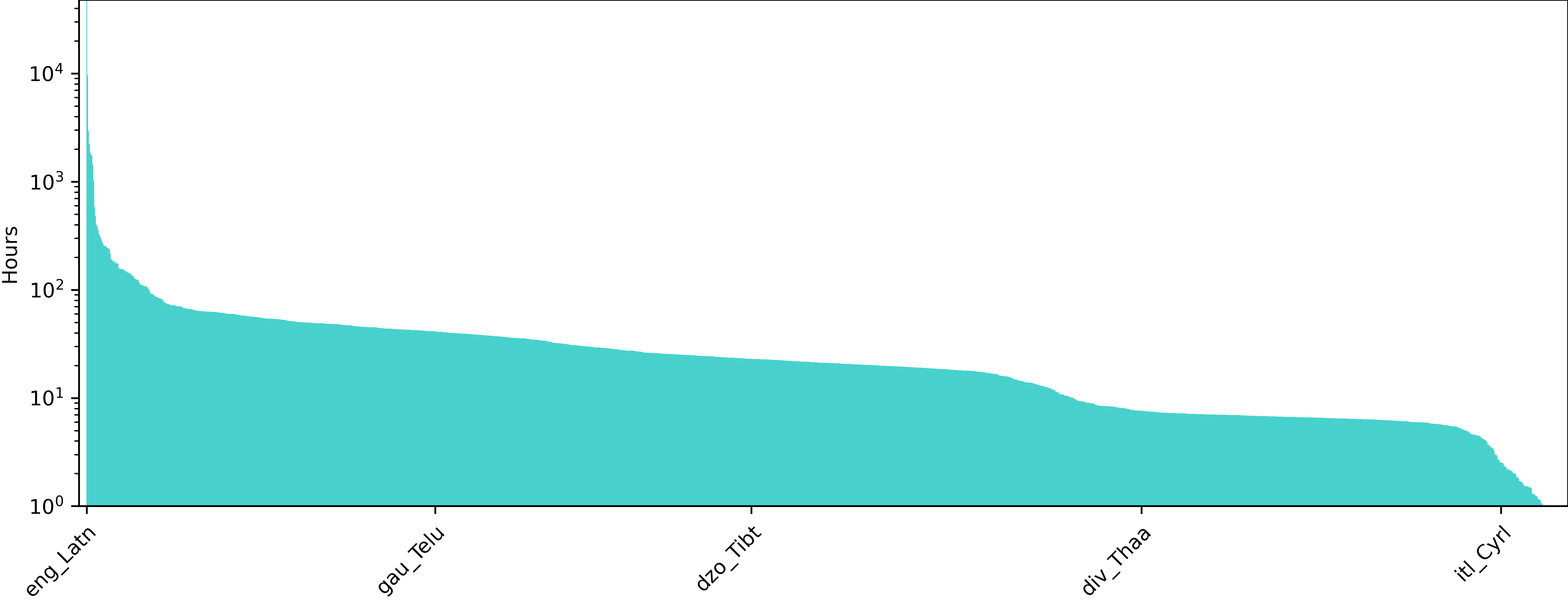}
  \caption{Statistics of the \allasr labeled data (hours of speech recordings paired with transcription) used to pre-train \OmniASR.}
  \label{fig:ft_train_stats}
\end{figure}

Once data preparation was complete, we combined all cleaned ASR datasets described in \Crefrange{sec:existing_asr_sources}{sec:vendor_corpus_desc} into a unified corpus, which we refer to as \allasr. Summary statistics of \allasr are shown in \Cref{tab:all_asr_dataset}, and its overall distribution is illustrated in \Cref{fig:ft_train_stats}. Beyond expanding language coverage, consolidating diverse ASR corpora into a single dataset improved model robustness to varied audio conditions, as demonstrated in \Cref{sec:corpus_holdout_ablations}.

\begin{figure}[ht]
  \centering
  \includegraphics[width=0.8\textwidth]{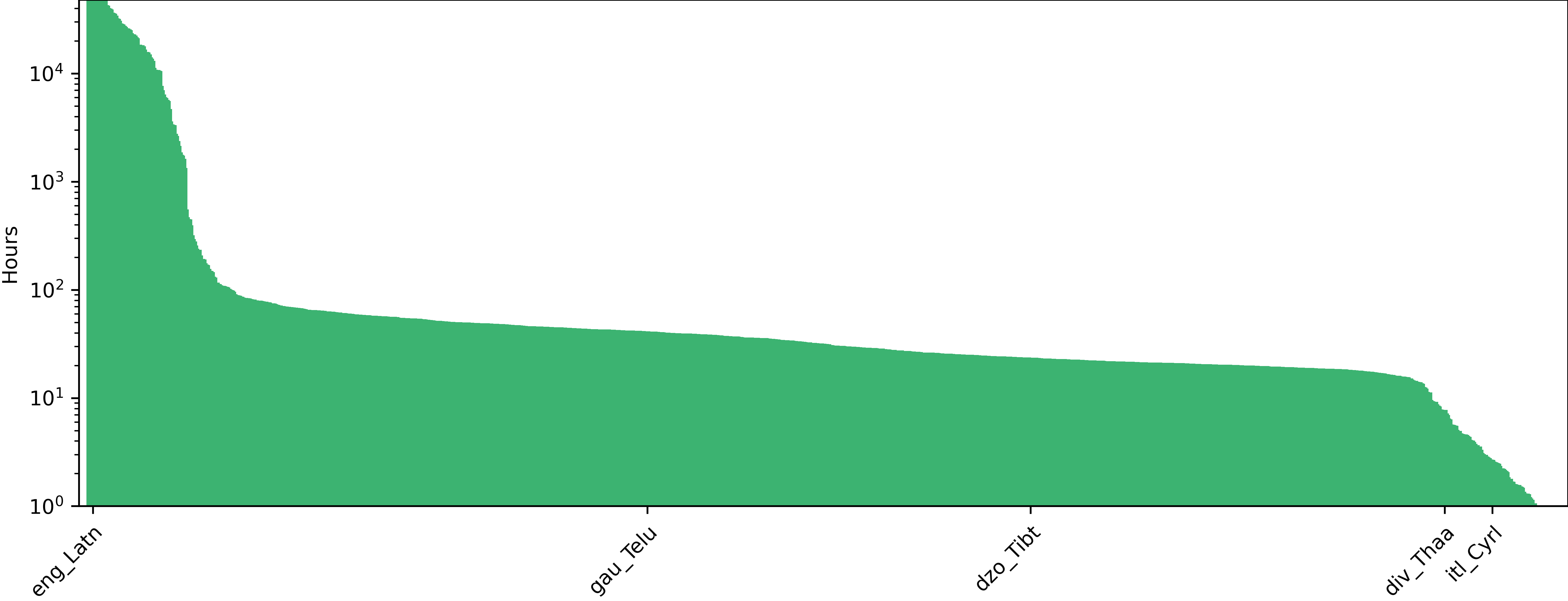}
  \caption{Statistics of the unlabeled data (hours of speech recordings) used to fine-tune \OmniASR for the ASR task.}
  \label{fig:pt_train_stats}
\end{figure}

In parallel, the unlabeled speech data described in \Cref{sec:ssl_data} was consolidated into a single corpus for self-supervised pre-training. Long recordings were segmented into chunks no longer than $30,\text{s}$ to standardize training inputs. The overall distribution of this dataset is shown in \Cref{fig:pt_train_stats}.

\begin{table}[!ht]
    \centering
    \begin{tabular}{p{5cm}p{5cm}p{5cm}}
    \toprule
         & Number of hours (total) & Number of languages \\
    \midrule
    Open source datasets & 15,000 & 200 \\
    LTPP, internal \& licensed data & 150,100 & 1,100\\
    \midrule
    African Next Voices & 7,200 & 13 \\
    Open Multilingual Speech Fund & 1,940 & 177 \\
    Lanfrica/Naijavoices & 110 & 11 \\
    \midrule
    \vendorcorpus & 3,350 & 348 \\
    \midrule
    \textbf{Total} & 120,710 & 1,690 \\
    \bottomrule
    \end{tabular}
    \caption{Summary statistics of the training split of the combined \allasr dataset.
    }
    \label{tab:all_asr_dataset}
\end{table}

Due to the heterogeneous nature of the datasets required to represent such a broad spectrum of languages---including variations in recording conditions, speaker demographics, and domain coverage---our development and test data splits are also necessarily heterogeneous. As a result, we caution readers against making direct comparisons between results obtained on different benchmarks. For example, error rates reported on MMS-lab, which features only a handful of speakers per language and contains high-quality recordings, are not directly comparable to those from more diverse datasets such as our own \vendorcorpus or the latest spontaneous speech data from Common Voice---which encompass a much wider range of speakers and recording conditions. This is further unpacked and demonstrated in \Cref{sec:omni_asr_corpus_holdout}.

\section{\OmniASR Models}
\label{section:models}

This section introduces the \OmniASR models. At a high level, all models follow an encoder–decoder architecture. The speech encoder is a large Transformer~\citep{vaswani2017attention} network that extracts high-level cross-lingual representations from input utterances, while the text decoder—either a linear layer or a Transformer decoder—maps these representations into character tokens.

We begin in \Cref{sec:w2v2} by describing how the speech encoder is developed to initialize with strong, massively multilingual speech representations. \Cref{sec:asr_methods} then details the creation of our ASR systems, covering both a traditional CTC-based approach and a novel LLM-based approach.

Even with the broad coverage of our supervised ASR models, some languages inevitably remain unsupported. To address this, \Cref{sec:zs_asr} introduces a zero-shot extension of our LLM-based models. We show that by providing only a few in-context examples at inference time, the models can perform ASR on previously unseen languages. \Cref{sec:zs_icl} further investigates strategies for selecting and constructing these in-context examples to maximize zero-shot performance.

Last but not least, we demonstrate the flexibility of our LLM-based ASR models by repurposing them for speech-to-text translation~(S2TT). Remarkably, this requires no dedicated S2TT optimization recipe or complex training pipeline, yet achieves strong performance compared to existing state-of-the-art systems. We detail these results in \Cref{sec:s2tt_results}.

\subsection{Massively Cross-Lingual Self-Supervised Representations}
\label{sec:w2v2}

At the core of \OmniASR is the speech encoder, whose quality directly determines ASR performance. To ensure that the encoder can extract high-level semantic representations across the wide range of languages we aim to cover, we adopted wav2vec2.0 \citep{baevski2020wav2vec} for self-supervised learning~(SSL), leveraging a large-scale corpus of unlabeled speech. We further scaled wav2vec~2.0 to increase model capacity, enabling it to capture massively multilingual speech representations. We then pre-trained a 7B-parameter wav2vec~2.0 model on 4.3M hours of speech, drawn from a combination of public and internal corpora spanning more than 1,600 languages. To our knowledge, this constitutes one of the largest publicly available SSL model to date, both in terms of parameter count and language coverage. The following sections describe in detail how this was achieved.

\subsubsection{Self-supervised Pre-training with wav2vec~2.0}
Although first proposed in~2020,~wav2vec~2.0~\citep{baevski2020wav2vec} remains one of the most prominent and effective algorithms for self-supervised learning of speech representations. The basic architecture of~wav2vec~2.0 consists of a convolutional feature encoder, a Transformer encoder network, and a quantization module. The convolutional feature encoder $f: \mathcal{X} \mapsto \mathcal{Z}$ maps raw audio~$\mathcal{X}$ to a latent representation $Z = (z_1, z_2, ..., z_T)$, where each~$z_t$ here corresponds to~25ms of audio strided by~20ms. The Transformer encoder $g: \mathcal{Z} \mapsto \mathcal{C}$ then processes $Z$ into contextualized representations $C = (c_1, c_2, \ldots, c_T)$. In parallel, the quantization module $h: \mathcal{Z} \mapsto \mathcal{Q}$ discretizes $Z$ into $Q = (q_1, q_2, \ldots, q_T)$, which are used as learning targets in the objective.

Training proceeds via solving a contrastive task over masked feature encoder output~$Z$. More specifically, spans of time steps in~$Z$ are randomly masked, and the objective requires identifying the true quantized latent~$q_t$ for a masked time step~$z_t$ within a set of distractors sampled from other masked time steps of the same utterance, denoted as $\tilde{q}\in Q$. The loss to minimize is defined as:
\begin{equation}
    -\log \frac{\exp(sim(c_t, q_t))}{\sum_{\tilde{q}\sim Q} \exp(sim(c_t, \tilde{q}))},
\end{equation}
where~$sim$ stands for cosine similarity, and~$Q$ includes~100 distractors and the ground truth~$q_t$ itself.  Once trained, the quantization module can be discarded, and only the convolutional feature encoder and the Transformer encoder network are required for downstream usage.

\subsubsection{Scaling Speech SSL Beyond 2B}
Beyond designing effective SSL objectives, model capacity is equally—if not more—crucial to improving representation quality. Since the release of the original 300M-parameter wav2vec2.0 model \citep{baevski2020wav2vec}, which at the time was considered large and demonstrated unprecedented success in speech SSL, researchers have pursued two parallel directions: refining SSL algorithms~\citep{hsu2021hubert,chen2022wavlm,chung2021w2v,chiu2022self} and scaling up model size to exploit the potential of ever-larger unlabeled corpora. To date, the largest publicly reported speech SSL models are Google’s Universal Speech Model (USM) \citep{zhang2023google} and Meta’s XLS-R \citep{babu2021xls}, both reaching approximately 2B parameters.

Yet it remains an open question whether 2B parameters marks the effective limit of scaling, either because additional capacity yields diminishing returns, or because 2B parameters are already sufficient for solving most speech tasks. In this work, we revisit the scaling laws of speech SSL by extending wav2vec2.0 from 300M to 1B, 3B, and ultimately 7B parameters. All models are trained on a collection of 4.3M hours of public and internal speech corpora covering more than 1,600 languages (see \Cref{sec:final_data}).

\textbf{Pre-training Setup}

\begin{table}[htbp]
  \centering
  \begin{tabular}{lrrrrr}
    \toprule
    Model & \# of layers & model dim & ffn dim & \# of attn heads & \# params \\
    \midrule
    OmniASR-W2V-0.3B   &  24  &  1024  &  4096  &  16  &  317M  \\
    OmniASR-W2V-1B     &  48  &  1280  &  5120  &  16  &  965M  \\
    OmniASR-W2V-3B     &  60  &  2048  &  8192  &  16  &  3046M \\
    OmniASR-W2V-7B     & 128  &  2048  &  8192  &  16  &  6488M \\
    \bottomrule
  \end{tabular}
  \caption{\OmniASR cross-lingual pre-trained wav2vec~2.0 models.}
  \label{tab:w2v2_config}
\end{table}

The configurations of our wav2vec2.0 models—including the 300M, 1B, 3B, and 7B variants—are summarized in \Cref{tab:w2v2_config}. We trained all models using the fairseq2 framework~\citep{balioglu2023fairseq2}. Because our pre-training data spans many languages and multiple sources, balancing across domains and languages was essential. To this end, we employed a two-step sampling procedure. First, for each data source, we sample the data for the $L$ different languages from a distribution
\begin{equation}
\label{eq:betas}
p_l \sim \left(\frac{n_l}{N}\right)^{\beta_{L}},
\end{equation}
where $l = 1, . . . , L$,~$n_l$ is the amount of unlabeled audio for each language in the current data source,~$N$ is the total amount of unlabeled audio in the current data source, and~$\beta_L$ is the upsampling factor which controls the trade-off between high- and low-resource languages during pre-training.  Second, we balanced the different data sources by treating each source as a language and applying the same sampling scheme with a sampling parameter~$\beta_D$.  In practice, we set both~$\beta_L$ and~$\beta_D$ to~0.5.  

All our pre-trained models were optimized with Adam~\citep{kingma2014adam} with a learning rate of~$1e-4$, which was warmed up for the first~32K steps followed by polynomial decay to zero for the remainder of training for a total of one million updates. Training batch sizes (measured in hours of audio per batch) were 6, 5.7, 8.5, and 17.6 for the 300M, 1B, 3B, and 7B models, respectively.

\subsection{Automatic Speech Recognition}
\label{sec:asr_methods}

We built on top of the wav2vec2.0 speech encoders described in \Cref{sec:w2v2} to construct two variants of ASR models. The first variant is a connectionist temporal classification (CTC)~\citep{graves2006connectionist} model, a framework designed to handle input and output sequences of varying lengths without requiring explicit alignments. CTC has become a foundational method in speech recognition and other temporal sequence tasks. By enabling models to learn alignments implicitly, CTC effectively captures temporal dependencies and has driven state-of-the-art performance in multiple applications. Our CTC models comprise of a single linear layer on top of a speech encoder. During training, the speech encoder was seeded from pre-trained wav2vec~2.0, and the entire model was optimized simultaneously using a CTC loss.

\begin{figure}[htbp]
  \centering
  \includegraphics[width=0.5\textwidth]{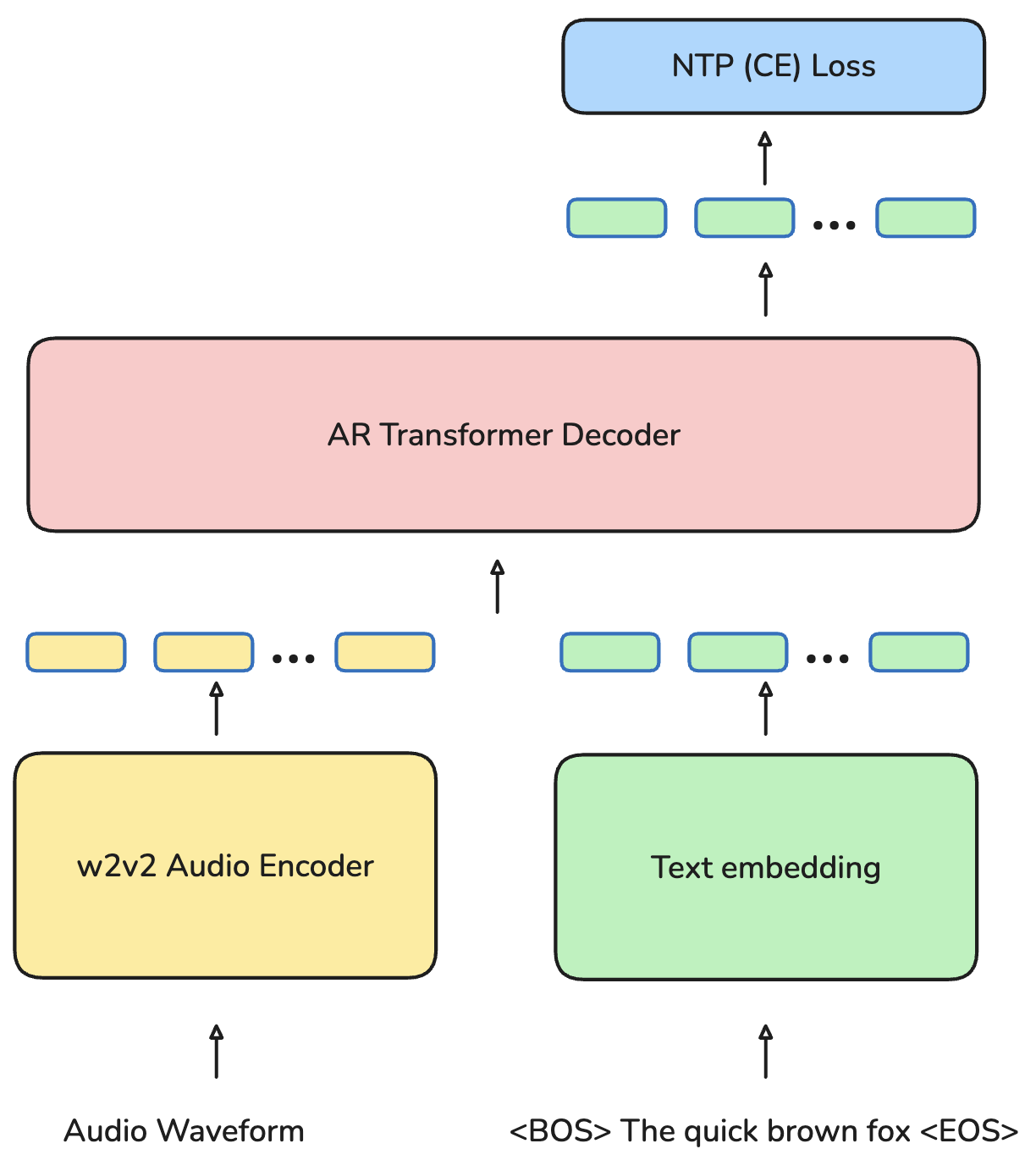}
  \caption{The LLM-ASR model architecture. A wav2vec~2.0 speech encoder and a text embedding matrix embed the speech and text modalities. An autoregressive Transformer decoder emits text tokens, and the system is trained with a next-token prediction objective.}
  \label{fig:llm_asr}
\end{figure}

Transformer decoders have achieved state-of-the-art performance in natural language processing tasks by effectively modeling complex sequential dependencies. In ASR, stacking a Transformer decoder on top of a speech encoder enables the system to leverage rich acoustic representations while capturing long-range context. This hybrid architecture combines the strengths of speech-specific encoders with the powerful contextual modeling capabilities of Transformers~\citep{baevski2021unsupervised,radford2023robust}. As a result, it improves transcription accuracy and robustness in diverse speech recognition scenarios. In the rest of the paper, we refer to this architecture as LLM-ASR, since it uses the same Transformer decoder module commonly found in LLMs. Our LLM-ASR model consists of a speech encoder initialized from a pre-trained wav2vec~2.0 encoder and a Transformer decoder on top of it. The LLM-ASR architecture is depicted in Figure \ref{fig:llm_asr}.

Formally, both ASR models process a speech segment $x$ through a waveform audio encoder $g_s$. We denote $y$ as the text transcription sequence corresponding to the speech segment. Our LLM-ASR model additionally holds a text embedding matrix $g_t$, which maps text tokens and special tokens to vector representations in the Transformer model dimension. The base version of our LLM-ASR model operates on sequences of the form
\[ g_s(x) \,\, g_t(\text{<BOS>}) \,\, g_t(y) \,\, g_t(\text{<EOS>}). \]
where $<$BOS$>$ and $<$EOS$>$ denote beginning- and end-of-sequence tokens. This model was then trained using a standard next-token prediction criterion (cross-entropy) to generate the transcription $y$ followed by an end-of-sequence token.

\subsection{Zero-Shot Speech Recognition for Unseen Languages}
\label{sec:zs_asr}
 
Our supervised ASR models described above support over 1,600 languages using labeled data. However, there remain languages for which no labeled data are available and which therefore cannot be supported by this purely supervised approach. To address this gap, we extend our LLM-ASR model with a zero-shot capability that allows it to perform ASR in any language or domain—including those unseen during training.

\begin{figure}[htbp]
  \centering
  \includegraphics[width=1.0\textwidth]{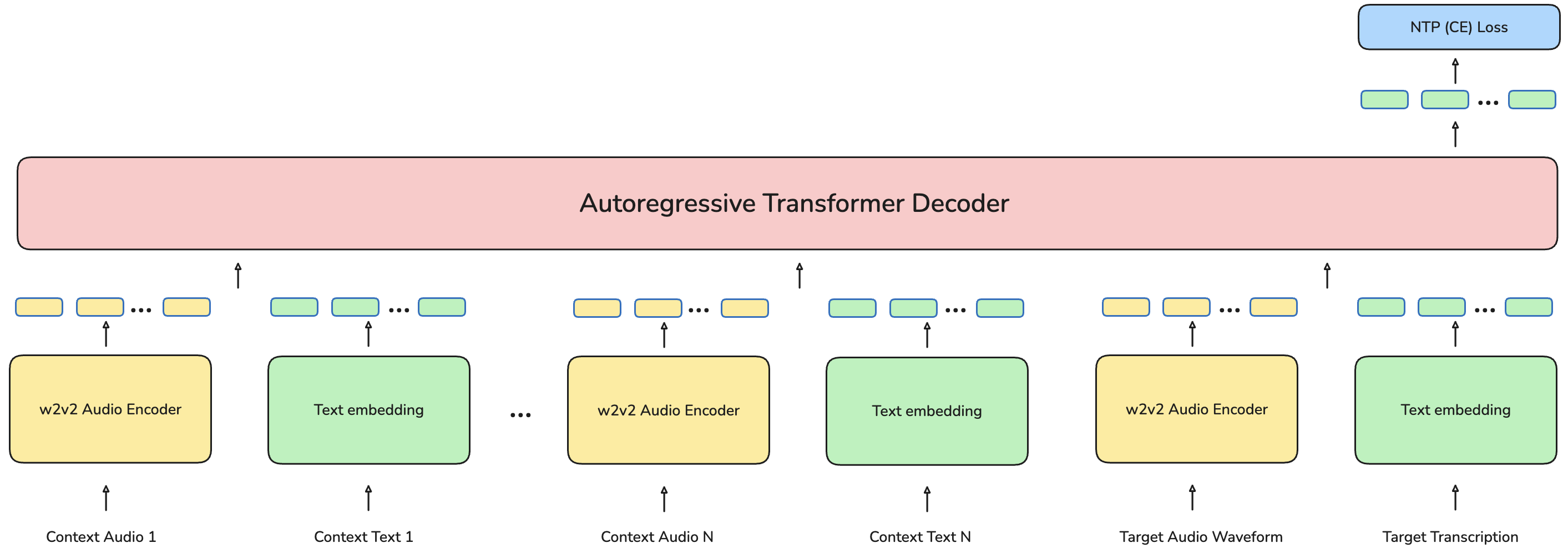}
  \caption{The LLM-ASR model architecture with context examples. Special tokens are omitted for simplicity.}
  \label{fig:context}
\end{figure}

The key idea is to shift from single-sample supervision to context-based training. At training time, instead of providing the model with only one speech–text pair, we present $N+1$ pairs from the same language. The first $N$ pairs serve as context examples and are prepended to the Transformer decoder prompt. The final pair is the target sample, whose transcription the model is trained to predict in the standard next-token prediction framework. This design teaches the model to condition on a few examples of speech–text pairs from a language before producing a transcription for a new utterance in the same language. Because our training corpus covers a large number of languages, we hypothesize that this behavior generalizes to languages absent from training data. As a result, the model acquires a zero-shot ASR capability, effectively enabling communities to extend recognition to their own languages with only a handful of paired examples. The overall architecture of the zero-shot model is illustrated in \Cref{fig:context}.

In technical terms, we denote the additional $N$ context speech-text pairs as $\left(x^c_i, \, y^c_i\right)$, where $i \in \{1, ..., N\}$. Each pair is then embedded with the appropriate modality encoder for the speech and text parts: $g_s\left(x^c_i\right),  \, g_t\left(x^c_i\right)$. The Transformer decoder then operates on the following sequence syntax:

\[ \text{<c> \{<cs>} \,\, g_s(x^c_i) \,\, \text{<cs BOS>} \,\, g_t(x^c_i) \,\, \text{<cs EOS> </cs>\}} \times \text{N </c>} \,\, g_s(x) \,\, \text{<BOS>} \,\, g_t(y) \,\, \text{<EOS>}, \]

where <c>, </c>, <cs>, </cs>, <cs BOS> and <cs EOS> are special tokens denoting the beginning and end of the context, each context example, and the text part within a context example. Each special token is embedded as a text token using $g_t$, which is omitted above for simplicity of notation. The model was then trained to predict $g_t(y)$ and the final <EOS> using the standard next-token prediction objective. The above sequence syntax, except the last $g_t(y)$ and <EOS>, is referred to as the model prompt. At inference time, this prompt is provided and the model generates a candidate transcription $\hat{y}$ and <EOS>.

\subsection{Selection of Context Examples for Zero-Shot ASR}
\label{sec:zs_icl}

In \Cref{sec:zs_asr}, we showed that zero-shot ASR can be performed by providing a few context examples from the target language. At inference time, we have the flexibility to choose which examples to provide, and different construction strategies can significantly impact model performance.
Formally, given a target utterance (the query) and a set of candidate speech–transcription pairs (the retrieval base), the task is to select the context examples that maximize transcription accuracy.

As a baseline, examples can be chosen at random within the target language. To improve over this, one natural strategy is to retrieve context examples that are acoustically or semantically similar to the target utterance. A straightforward approach is to embed the target audio into a fixed-length vector and perform nearest-neighbor search within the retrieval base. Prior work on Whisper has shown that kNN-based example selection can improve in-context ASR performance~\citep{wang2024can}.

For our work, we leverage the SONAR encoder~\citep{duquenne2023sonar} as the embedding model to retrieve context examples. SONAR is a multilingual and multimodal system capable of transforming audio or text into a fixed-sized sentence embedding with rich semantic information. In practice, we embedded the target audio sample and used it as the query, while the retrieval base was represented by embeddings of both speech and text. Context examples could then be selected based on nearest-neighbor similarity between the query embedding and the embeddings of the retrieval candidates.

\subsection{Conditioning on Language Codes} \label{sec:lid}

Multilingual ASR models generally demonstrate the ability to detect the spoken language implicitly and transcribe it correctly~\citep{pratap2024scaling,radford2023robust}. However, our initial experiments revealed some limitations to this ability. For example, certain languages such as Urdu can be written in multiple scripts, which creates ambiguity for the model. In other cases, closely related languages in the training set may confuse the model about which language to use for transcription. Moreover, in many real-world applications, the user already knows the spoken language in advance and would benefit from being able to provide this information explicitly.

To address these issues, we introduce a mechanism for supplying the model with an additional optional input: a language code together with the desired script. This information is encoded using a dedicated embedding matrix. Specifically, we assign each observed combination of language and script in the training corpus a unique ID, reserving ID 0 to denote an unknown language. During training, this ID—denoted $l$—is embedded through a matrix $g_l$. The input sequence to the model becomes
\[ g_s(x) \,\, g_t(\text{<language>}) \,\, g_l(l) \,\,g_t(\text{<BOS>}) \,\, g_t(y) \,\, g_t(\text{<EOS>}), \]
where <language> is a newly introduced special token. To ensure the model can function both with and without explicit language information, we randomly drop the language input during training with probability $p$. This enables flexible inference modes: either conditioned on a known language and script or left unconstrained when no prior information is available.

\section{Model Training and Evaluation}
\label{section:eval}

In this section, we present the training details of \OmniASR models and outline the extensive experiments to validate their capabilities. We begin with the traditional supervised setting in \Cref{sec:main_results} and~\ref{sec:eval_1600}. First, we compare \OmniASR with existing large-scale multilingual ASR systems, including Whisperv3 from OpenAI \citep{radford2023robust}, the Universal Speech Model (USM) from Google~\citep{zhang2023google}, and Massively Multilingual Speech (MMS) from Meta~\citep{pratap2024scaling}, and demonstrate our state-of-the-art performance on languages overlapping with these existing multilingual systems. We then analyze performance across the full set of 1,600+ supported languages, including more than 500 never before covered by any ASR system.

To extend \OmniASR’s capabilities to support virtually any spoken language, we previously introduced our zero-shot model in \Cref{sec:zs_asr}. In \Cref{sec:zs_asr_results} and~\ref{sec:sonar_results}, we show that this model successfully transcribes utterances from languages entirely unseen during training. In \Cref{sec:s2tt_results}, we further adapt the LLM-ASR variant to perform speech-to-text translation with minimal modification, requiring only the insertion of source and target language identifier (LID) tokens into the input sequence. Finally, we present an ablation study on fine-tuning data-mixing (\Cref{sec:datamix_ablations}) and an analysis of the impact of conditioning on language codes (\Cref{sec:lid_results}).

\subsection{ASR Training Setup}
\label{sec:asr_training_setup}
We trained multilingual ASR models by fine-tuning the pre-trained SSL speech encoders introduced in \Cref{sec:w2v2} using the labeled data described in \Cref{sec:final_data}. For both CTC and LLM-ASR models, we consider four encoder sizes: 300M, 1B, 3B, and 7B parameters. All LLM-ASR variants use the same decoder configuration: a 12-layer Transformer with model dim 4096 and eight attention heads, totaling 1.2B parameters. Throughout, we refer to the LLM-ASR variants by their encoder size. 

\textbf{CTC optimization details.} To emit transcriptions, we added a linear layer on top of the pre-trained SSL models, which maps their output to a vocabulary consisting of the set of characters appearing in our labeled training corpus for all languages. We then fine-tuned the entire network with the connectionist temporal classification (CTC) criterion \citep{graves2006connectionist}. We used Adam~\citep{kingma2014adam} with exponential decay rates~$\beta_1 = 0.9$, $\beta_2 = 0.98$ to optimize model parameters using a tri-stage schedule: warm-up over the first 10\% of updates, hold constant for the next 40\%, and exponential decay for the final 50\%. All CTC models were trained with a learning rate of $10^{-5}$, an effective batch size of ~4.2 hours, and for 200k steps.

\textbf{LLM-ASR optimization details.} The LLM-ASR models introduced in \Cref{sec:asr_methods} were trained with the same character set described above under a next-token prediction (cross-entropy) objective. Adam was used for those models as well, with a learning rate of $5 \times 10^{-5}$ and the same~$\beta$ values and learning rate schedule as above. The effective batch size of those models was set to~2.1 hours and the model was trained for~150k steps. At inference time, our LLM-ASR models use beam search decoding with a beam size of five hypotheses.

\subsection{Comparison to Other Work}
\label{sec:main_results}

Below, we compare \OmniASR to some of the most prominent existing multilingual ASR work, including Whisper~\citep{radford2023robust},  Universal Speech Model~(USM)~\citep{zhang2023google}, and Massively Multilingual Speech~(MMS)~\citep{pratap2024scaling}.

\subsubsection{\OmniASR vs. Whisper}

\begin{table}[htbp]
  \centering
  \begin{tabular}{lrrrrrrrr|cc}
    \toprule
    Model & \multicolumn{2}{c}{MMS-Lab-66} & \multicolumn{2}{c}{FLEURS-81} & \multicolumn{2}{c}{MLS-8} & \multicolumn{2}{c}{CV22-76} & \multicolumn{2}{c}{Win Rate} \\
    & dev & test & dev & test & dev & test & dev & test &  $n=81$ & \makecell{$n=34$\\(top 50)} \\
    \midrule
    \textit{Prior Work}  \\
    Whisper small       &  66.8 & 	64.3 & 	51.5 & 	50.8 & 	6.2 & 	4.9 & 	103.6 & 	111.7  &  -  &  -  \\
    Whisper medium      &  55.5 & 	54.5 & 	48.0 & 	47.8 & 	6.8 & 	4.6 & 	79.8 & 	87.9  &  -  &  -  
    \\
    Whisper large-v3    &  32.0 & 	30.9 & 	22.0 & 	22.6 & 	2.3 & 	2.0 & 	27.3 & 	55.6   &  -  &  -  \\
    \midrule
    \textit{This Work}  \\
    300M CTC            &   4.9  &   4.7  &  11.7  &  11.8  &  4.6  &  4.1  &  16.7  &  17.6  & 37  &  -  \\
    1B CTC              &   3.0  &   2.8  &   8.5  &   8.6  &  3.3  &  3.1  &  13.5  &  14.8  & 48  &  -  \\
    3B CTC              &   2.2  &   2.0  &   7.7  &   7.8  &  3.1  &  2.7  &  12.3  &  13.7  & 54  &  -  \\
    7B CTC              &   1.9  &   1.7  &   7.2  &   7.3  &  2.8  &  2.5  &  11.6  &  13.8  & 61  &  -  \\
    300M LLM-ASR        &   1.7  &   1.9  &   8.0  &   7.8  &  3.6  &  3.2  &   6.5  &   7.1  & 46  &  -  \\
    1B LLM-ASR          &   1.4  &   1.2  &   6.7  &   6.6  &  2.9  &  2.7  &   5.9  &   6.5  & 55  &  -  \\
    3B LLM-ASR          &   1.3  &   1.1  &   6.3  &   6.2  &  2.8  &  2.6  &   6.3  &   6.6  & 57  &  -  \\
    7B LLM-ASR          &   1.1  &   1.0  &   5.9  &   5.6  &  2.5  &  2.4  &   5.5  &   6.4  & 65  &  24 \\
    7B LLM-ASR + LM     &    -   &    -   &   5.7  &   5.5  &  2.5  &  2.4  &    -   &    -   & 65  &  -  \\
    \bottomrule
  \end{tabular}
  \caption{Comparison against Whisper v3, including its large~(1.5B), medium~(769M), and small~(244M) variants.  For each benchmark, we report average CER across languages on both dev and test splits. The comparison only considers languages that Whisper covers in each benchmark, and the number that follows the dataset name indicates the number of languages considered. The two rightmost columns show the win rate of our model against Whisper large v3 on the FLEURS test set: $n=81$ considers the entire FLEURS-81 languages, while $n=34$ only considers the top~50 most spoken languages in the world that are covered by FLEURS (34 of them).}
  \label{tab:vs_whisper}
\end{table}

Whisper is a multilingual speech model trained on approximately 5M hours of weakly labeled web audio and supports a range of speech-processing tasks, including ASR in 99 languages. Its architecture is a Transformer-based sequence-to-sequence model~\citep{sutskever2014sequence}, consisting of an encoder and a decoder, with the decoder functioning in part like a language model. Thanks to its strong performance and easily accessible API, Whisper has become one of the most widely adopted speech models in the research and developer communities.

In \Cref{tab:vs_whisper}, we compare \OmniASR models against Whisper’s latest large-v3 release, as well as its smaller variants, using the MMS-Lab~\citep{pratap2024scaling}, FLEURS~\citep{conneau2023fleurs}, MLS~\citep{pratap2020mls}, and Common Voice 22 (CV22)~\citep{ardila2020common} evaluation sets. We report character error rate (CER) averaged across languages. In this comparison, we only considered languages that Whisper covers in each benchmark; the number following each dataset name indicates the corresponding number of languages evaluated. To further strengthen the comparison, we also trained n-gram language models for FLEURS and MLS languages using their training transcripts, and considered LM fusion with those models for our largest variant using hyperparameters optimized on the dev set. The main results from this comparison are summarized in \Cref{tab:vs_whisper}.

More specifically, we find that even our smallest model outperforms Whisper large-v3 on most evaluation sets, as measured by average CER across languages. Our 300M-CTC variant surpasses Whisper-large on MMS-Lab-63, FLEURS-82, and CV22-76, and falls behind only on MLS-8. As we scale encoder size, the gap with Whisper on the former three benchmarks continues to widen. Against Whisper small and medium, the 300M-CTC outperforms them on all four benchmarks.

Moreover, \OmniASR performs strongly on the world’s most spoken languages while supporting long-tail ones. Whisper shows strength on some of the highest-resource languages, as reflected in its MLS-8 results, likely due to the large amount of labeled training data in those languages. However, its accuracy drops sharply on long-tail languages included in other benchmarks. Our models, on the other hand, while remaining strong on high-resource languages, outperform significantly on long-tail languages. In general, we find that the Whisper models' average CER across languages is disproportionately affected by a long set of poorly supported languages. To provide additional insights to the comparisons, \Cref{tab:vs_whisper} reports the number of languages on which our models outperform Whisper large-v3 on FLEURS-81, including a breakdown for the 34 of the world’s 50 most spoken languages%
\footnote{\url{https://www.ethnologue.com/insights/ethnologue200}}
that are covered in FLEURS-81.  Comparing our 7B-LLM against Whisper large-v3, we achieve an 80\% win rate (65 out of 81) across all languages in FLEURS-81, and 71\% (24 out of 34) on the most spoken languages.

Finally, comparing our own variants, the LLM models consistently outperform their CTC counterparts by a wide margin. Error analysis shows that CTC models often fail due to script misprediction: when the wrong script is chosen for an input utterance, the decoded characters belong to another language altogether. This issue is particularly common in low-resource settings as models are less familiar with their scripts. By contrast, our LLM-ASR models benefit from the ability to condition on language codes at inference time (while still working without them), which largely resolves the wrong-script problem. The LLM results in \Cref{tab:vs_whisper} are reported with language conditioning. Ablations on language conditioning are presented in \Cref{sec:lid_results}.

\subsubsection{\OmniASR vs. USM}
USM and \OmniASR follow a broadly similar development recipe: both begin with large-scale self-supervised pre-training of a Transformer encoder, followed by appending a decoder on top and fine-tuning the entire model with labeled data. In USM’s case, the encoder adopts a Conformer architecture~\citep{gulati2020conformer}, a convolution-augmented Transformer variant. Pre-training is performed with the BEST-RQ algorithm~\citep{chiu2022self} on roughly 12M hours of proprietary YouTube audio spanning 300 languages, and fine-tuning for ASR is carried out on 90K hours of labeled data across 100 languages. The Conformer encoder itself has 2B parameters, and the decoder is an RNN-Transducer that has a built-in neural language model. Additional USM variants (e.g., USM-M and USM-M-adapter) extend this setup with multi-stage pre-training pipelines that include text pre-training and labeled audio, totaling about 20K hours. In contrast, \OmniASR encoders are pre-trained solely on unlabeled speech data.

\begin{table}[htbp]
  \centering
  \begin{tabular}{lrrr}
    \toprule
    Model & \multicolumn{2}{c}{FLEURS-102}  \\
    & dev & test  \\
    \midrule
    \textit{Prior Work}  \\
    Maestro-U~\citep{chen2022maestro}  &   -   &  8.7  \\
    USM                                &   -   &  6.9  \\
    USM-M                              &   -   &  6.5  \\
    USM-M-adapter                      &   -   &  6.7  \\
    \midrule
    \textit{This Work}  \\
    7B CTC                             &  7.4  &  7.5  \\
    1B LLM-ASR                         &  7.3  &  7.2  \\
    3B LLM-ASR                         &  6.8  &  6.7  \\
    7B LLM-ASR                         &  6.4  &  6.2  \\
    7B LLM-ASR + LM                    &  6.2  &  6.1  \\
    \bottomrule
  \end{tabular}
  \caption{Comparison against USM and its variants on FLEURS-102.  We report average CER across languages.  For USM and its variants, only test set results are available; we report our results on both dev and test splits.}
  \label{tab:vs_usm}
\end{table}

Since USM and its variants are not publicly accessible, we rely on their reported results on FLEURS-102, presented in \Cref{tab:vs_usm}. We see that when considering the full FLEURS-102 benchmark~(as opposed to FLEURS-81 in \Cref{tab:vs_whisper}), our 7B-LLM model still outperforms 7B-CTC. Compared to the best USM variant~(USM-M), which achieves a CER of~6.5\%, our 7B-LLM achieves~6.2\%, and when we incorporate LM fusion at inference, the CER is further reduced to~6.1\%.  Despite the fact that our models are pre-trained on more than~50\% less unlabeled speech data than USM~(4.3M vs. 12M hours) and do not adopt a sophisticated pre-training pipeline involving multiple stages (as USM does), our models still outperform the USM models. We largely attribute this to the impact of encoder size scaling.

\subsubsection{\OmniASR vs. MMS}
Similar to USM and \OmniASR, MMS \citep{pratap2024scaling} takes advantage of SSL to leverage large quantities of unlabeled speech data to pre-train a Transformer encoder so as to initialize it with rich cross-lingual speech representations, before appending a decoder and fine-tuning the entire model with labeled data. Specifically, MMS uses wav2vec~2.0~\citep{baevski2020wav2vec} to train a~1B Transformer encoder network, leveraging around~500k hours of unlabeled speech data and covering approximately~1400 languages. After appending a linear layer as a decoder to the pre-trained encoder, the entire model is fine-tuned with around~45k hours of labeled data to cover ASR for approximately~1100 languages using CTC.

For FLEURS-102, MMS incorporates a sophisticated fine-tuning pipeline to optimize its ASR performance—the Transformer encoder is modified with adapter modules~\citep{houlsby2019parameter}, where a different set of adapter weights is used for each language. Specifically, MMS has an adapter module augmented to every layer of its Transformer encoder, where the adapter is added after the last feed-forward block. Each adapter module consists of a LayerNorm layer, a downward linear projection, followed by a ReLU activation, and an upward linear projection.
After an initial fine-tuning stage across all languages, MMS performs a second stage of language-specific fine-tuning. In this step, the model introduces a randomly initialized linear layer that maps to the output vocabulary of a language, alongside a dedicated language-specific adapter. These additional parameters are then fine-tuned on the labeled data available for that language.

\begin{table}[htbp]
  \centering
  \begin{tabular}{lrrrr}
    \toprule
    Model & MMS-Lab-1143 & FLEURS-102 & MLS-8\\
    \midrule
    \textit{Prior Work} \\
    MMS - single-domain training + LM   &   -    &  6.4   &  8.7  \\
    MMS - multi-domain training + LM    &  2.1   &  6.3   &  9.0  \\
    \midrule
    \textit{This Work}  \\
    7B LLM-ASR                          &  1.9   &  6.2   &  8.0  \\
    7B LLM-ASR + LM                     &   -    &  6.1   &  8.0  \\
    \bottomrule
  \end{tabular}
  \caption{Comparison against MMS on the test sets of MMS-Lab-1143, FLEURS-102, and MLS-8. We report average CER across languages except for MLS-8, where we report WER. ``MMS - single-domain training'' means the MMS model is fine-tuned on just that particular dataset, and ``MMS - multi-domain training'' means the model is fine-tuned on the full~45k hours of MMS labeled data. Both reported MMS results are with n-grams LM decoding.}
  \label{tab:vs_mms}
\end{table}

We compare MMS with \OmniASR in \Cref{tab:vs_mms}, reporting CER on MMS-Lab-1143 and FLEURS-102, and WER on MLS-8. The results are averaged across all the languages in the corresponding datasets. ``MMS - single-domain training'' means that the MMS model is fine-tuned on just that particular dataset, while ``MMS - multi-domain training'' means the MMS model is trained on their entire~45k hours of labeled data. After training, during inference time, MMS uses an n-gram model trained on Common Crawl for better decoding results.  From the table, we see that our 7B-LLM outperforms MMS on all evaluation sets, regardless of the setting for which MMS models are optimized.

\subsection{Evaluation on 1600+ languages}
\label{sec:eval_1600}

In the previous section, we compared \OmniASR with Whisper, USM, and MMS, showing that our models set or match state-of-the-art performance across existing multilingual benchmarks. We now turn to a broader analysis of \OmniASR’s performance on the full set of 1,600+ languages it supports—including more than 500 languages that have never before been covered by any ASR system.

Evaluating models at this scale requires a structured approaches. As such, we adopted two complementary protocols: (i) dividing languages into high-, mid-, and low-resource categories based on the amount of labeled training data available, and (ii) sorting languages into 14 major groupings following the principles outlined below. For simplicity, all test splits are aggregated by averaging results across languages within each category of the respective evaluation protocol.

\subsubsection{Evaluation based on Resource Buckets}

\begin{table}[htbp]
  \centering
  \begin{tabular}{lrrr}
    \toprule
     &  High  &  Mid  &  Low  \\
    \# of lang in this bucket   &  249 & 881  & 546  \\
    \midrule
    7B-CTC & $3.7\pm 0.7$ & $4.4\pm 0.6$ & $18.6\pm 1.2$ \\
    7B-LLM & $3.13\pm 0.7$ & $3.0\pm 0.3$ & $18.0\pm 1.2$ \\
    \bottomrule
  \end{tabular}
  \caption{Mean CER for each language-resource bucket with 95\% Confidence Intervals. High-resource languages have >50 hours training data, mid-resource have between 10-50h, and low- have <10h. Both models do not employ LM fusion.}
  \label{tab:resource_bucket_mean_cers}
\end{table}

\begin{table}[htbp]
  \centering
  \begin{tabular}{lrrr}
    \toprule
     &  High  &  Mid  &  Low  \\
    \# of lang in this bucket   &  249 & 881  & 546  \\
    \midrule
    7B-CTC & 231 & 823 & 184 \\
    7B-LLM & 236 & 841 & 195 \\
    \bottomrule
  \end{tabular}
  \caption{Number of languages within each resource-bucket where our models obtain CERs below~10.}
  \label{tab:resource_bucket_cer_threshold}
\end{table}

We group languages into resource buckets according to the amount of labeled training data available in \allasr{}. High-resource languages are those with more than 50 hours of training data, mid-resource languages fall between 10–50 hours, and low-resource languages have fewer than 10 hours. This results in 249, 881, and 549 languages in the high-, mid-, and low-resource buckets, respectively. To ensure a sufficient validation signal, we exclude languages with less than 30 minutes of data in their validation splits.

\Cref{tab:resource_bucket_mean_cers} reports the mean CER across languages in each bucket, while \Cref{tab:resource_bucket_cer_threshold} shows the number of languages achieving CER < 10 within each bucket. Both of our models can achieve low CERs (under 5) in the high- and mid-resource categories, with ~90\% of languages in these buckets meeting this threshold. On the low-resource bucket, where we have less than~10 hours of training data per language, the percentages of languages that meet the CER quality threshold fall to~34\% and~36\%, with an average CER of~18.6 and~18.0 for 7B-CTC and 7B-LLM, respectively. In \Cref{sec:language_specific_ablations}, we examine the performance of long-tailed languages and provide a recipe for further fine-tuning our models on specific languages to achieve optimal performance.

\subsubsection{Evaluation based on Language Groupings}
\begin{table}[htbp]
  \centering
  \begin{tabular}{lrrrr}
    \toprule
    Grouping & \# of lang & Avg CER & CER $\leq 10$ & \% \\
    \midrule
    Afroasia   &   92  &  11.8  &  61  &  66\%  \\
    Amazbasi   &   83  &   2.0  &  82  &  99\%  \\
    Amerande   &   67  &   2.0  &  66  &  99\%  \\
    Atlacong   &  389  &   9.3  & 280  &  72\%  \\
    Austasia   &   35  &   5.4  &  31  &  89\%  \\
    Austrone   &  239  &   5.1  & 193  &  81\%  \\
    Caucasus   &   35  &   3.9  &  35  & 100\%  \\
    Dravidia   &   22  &   7.3  &  18  &  82\%  \\
    Indoeuro   &  209  &   9.1  & 154  &  74\%  \\
    Mesoamer   &  159  &   7.8  & 115  &  72\%  \\
    Newguine   &   77  &   5.5  &  63  &  82\%  \\
    Nilosaha   &   56  &   4.4  &  50  &  89\%  \\
    Norameri   &   42  &   4.8  &  37  &  88\%  \\
    Sinotibe   &   65  &   8.2  &  52  &  80\%  \\
    \midrule
    Total      & 1570  &   7.1  & 1237 &  78\%  \\ 
    \bottomrule
  \end{tabular}
  \caption{Average CER across languages under~14 language groupings using our 7B-LLM model without LM fusion. We only considered languages that can be classified into one of the~14 groupings and dropped the rest of the languages our models support. \# of lang denotes the number of languages belonging to that particular grouping covered in our evaluation sets. CER~$\leq$~10 indicates the number of languages belonging to that grouping that achieves a CER no greater than~10, and~\% shows the percentage of that.}
  \label{tab:language_bucket}
\end{table}

The main principles used for grouping are as follows. Languages are first grouped according to their respective families; the definition of the term \textit{family} follows the linguistic genealogy research in \citet{hammerstrom2024glottolog}. In cases where family-based grouping does not yield a large enough number of group members (i.e., for either small families or families with a small number of members being represented in our datasets, as well as for language isolates), languages are additionally grouped by linguistic proximity. Although the eight-letter labels used for those groups (e.g., Caucasus, Norameri, Amerande) may sound geographical, linguistic proximity is not to be understood solely as geographical proximity but also as typological proximity (i.e., following aspects of linguistic typology). The grouping resulted in~14 groups of different sizes, ranging from~389 members for the largest group to~22 members for the smallest one.

In Table~\ref{tab:language_bucket}, we present the results of our 7B-LLM model across the~14 language groupings.  We omit languages our models support but cannot be classified into one of the~14 groupings in this analysis. \# of lang' denotes the number of languages under that particular grouping that are covered in our evaluation sets, and Avg CER shows the average CER across languages under that grouping. Additionally, in order to get a broader sense of quality, we measure the number of languages for which CER~$\leq$~10.  This indicates how many languages the model produces, on average, no more than one error in ten characters.  While this measure is very coarse, it enables us to get a sense of quality across such a large number of languages.  From the table, we see that overall our model meets the CER quality threshold for~78\% of the~1570 languages we evaluate on, and is able to reach a CER below~10 for all groupings except for Afroasia, for which we get~11.8.

By measuring our model's performance through the lens of resource buckets and language groupings, our analysis in \Cref{sec:eval_1600} demonstrates our models' ability to transcribe a massive variety of languages while maintaining reasonable to high quality.

\subsection{Accuracy of Zero-Shot Models on Unseen Languages}
\label{sec:zs_asr_results}

We conducted experiments to evaluate the generalization of our zero-shot ASR model described in \Cref{sec:zs_asr} to unseen languages. To that end, we excluded a set of 32 languages from our training set, which will be used for evaluation. The set of evaluation languages was chosen at random but in a manner that asserts that half of the languages are high-resource languages that are represented in more than one evaluation set, and the other half are low-resource languages that may only appear in a single evaluation set. Since some evaluation sets contain only a small number of the evaluation languages, it does not make sense to report accuracy by evaluation set in this setting. Instead, for each evaluation language, we compute its overall CER across all evaluation sets, and average this number across languages. The context examples were chosen randomly for each utterance from the same dataset and in a consistent manner across models. 

The zero-shot models are compared to a CTC and LLM-ASR baselines, both trained excluding the same set of languages, which are then used for evaluation. To find an optimal setting for generalizing to unseen languages, we experimented with a number of variants of the zero-shot model. The candidates vary by the number of context examples used, the seed used to initialize the speech encoder, and whether the speech encoder was frozen during that training or not. Results appear in \Cref{tab:fs}. From the table, we see that among baselines, the CTC model generalizes better to unseen languages than the LLM-ASR variant. However, when augmented with conditioning on context examples, the LLM-ASR model outperforms the CTC model and reduces the overall CER on unseen languages from 26.33\% to 14.4\% using a context size of 10, the largest context size we experimented with. Among zero-shot models, we found that seeding from CTC reduces the generalization ability to unseen languages. We also observed that tuning the speech encoder was crucial for demonstrating the zero-shot ability in a manner superior to baseline models.

An additional observation is that zero-shot models somewhat degrade accuracy on some datasets of seen languages compared to their non zero-shot counterparts. However, we release separate models for stronger support in the languages appearing in our training set, making this metric less important for zero-shot models. Two exceptions are the FLEURS-102 and CV22 datasets, in which zero-shot models outperform the baseline models. The reason for this is a relatively high number of utterances in those datasets where the script is being misrecognized by non zero-shot models, thus vastly increasing the CER. As zero-shot models are provided with a number of context speech and transcription pairs from the language, they significantly reduce script and language confusion errors. 

\begin{figure}[htbp]
  \centering
  \includegraphics[width=0.4\textwidth]{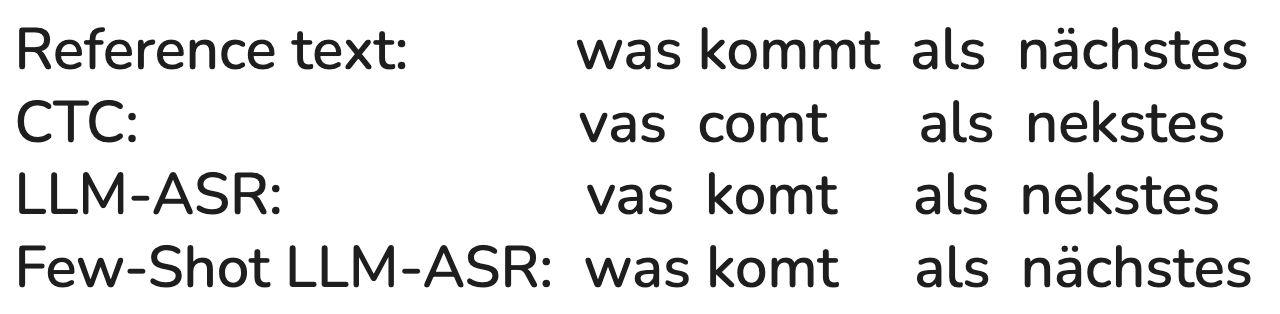}
  \caption{A German example of the zero-shot model (German was excluded from training of this model). While baseline models struggle with the correct spelling, the zero-shot ASR model produces a more accurate hypothesis.}
  \label{fig:zs_example}
\end{figure}

One example of the superiority of zero-shot models on unseen languages can be seen in \Cref{fig:zs_example}. This illustrates an example in German, which was excluded from training in all models in this subsection. While non zero-shot models make considerable spelling errors, zero-shot models do visibly better.  

\begin{table}[htbp]
\centering
\begin{tabular}{lccccccccc}
\toprule
Model & Context & Unseen & MMS-Lab & Omnilingual  & FLEURS & MLS & CV22 \\
      &         &        &         & ASR Corpus   & -102   &     &    \\
\midrule
CTC                         & 0 &  26.3  &  4.2  &	23.1  &	 8.5  &	 2.7  &  15.4 \\
LLM-ASR                     & 0 &  31.0	 &  2.9  &	20.3  &	 7.6  &	 2.7  &	 15.5 \\
ZS LLM-ASR, CTC seed        & 5 &  19.3	 &  3.4  &	21.2  &	 6.8  &	 2.9  &	  9.2 \\
ZS LLM-ASR, CTC seed, Fr.   & 5 &  26.5	 &  4.0  &	23.2  &	 8.0  &	 2.7  &	 11.8 \\
ZS LLM-ASR, w2v2 seed       & 5 &  17.6	 &  3.7  &	21.9  &	 7.1  &	 2.9  &	  8.7 \\
ZS LLM-ASR, w2v2 seed       & 10 & 14.4  &  4.3  &	23.2  &	 8.3  &	 3.1  &	 10.3 \\
\bottomrule
\end{tabular}
\caption{Generalization to unseen languages of the zero-shot models. Unseen refers to the language average CER across all evaluation sets for unseen languages. The rest of the evaluation sets specified refer to the portion of those sets with languages seen during training.}
\label{tab:fs}
\end{table}

\subsection{Constructing Context Examples for Zero-Shot ASR}
\label{sec:sonar_results}

In this section, we present a series of selection approaches for studying how the model uses context in the zero-shot ASR setting. Limited by the language coverage of the SONAR speech encoder, we trained another LLM-ASR with five context examples but with a different set of 32 holdout languages (supported by SONAR). We did not condition on language codes for this setting. The holdout languages remain diverse, encompassing languages with distinct scripts and belonging to various language groupings. Our set of holdout languages includes some very high resource languages, such as English and Spanish; most of the languages are mid-resource, ranging from 100-300 hours in the entire training corpora, and also a few lower resource languages below 100 hours, such as Welsh and Marathi. The model architecture and training basically follow \Cref{sec:zs_asr_results}. We initialized the speech encoder with the 7B wav2vec~2.0 encoder, and the speech encoder was updated during ASR training. After training, we evaluated zero-shot ASR performance on the holdout languages. For each evaluation set, we selected context examples from the corresponding training set for all selection approaches.

Intuitively, one strategy is to provide context examples that share similarities with the target; another is to sample a diverse set of context examples, where we try to cover as much variety of the unseen language as possible. An open question is which features to use when selecting context examples—textual, semantic, or audio similarity. These features are not entirely independent (e.g., higher semantic similarity can also lead to higher text overlap). In this section, the baseline approach would be randomly selecting context examples from the retrieval base without duplicates, and the random baseline, to some extent, would consist of diverse context examples of different aspects.

For selecting context examples that are similar to the target, we focused on these three features: text, semantic, and audio. For semantic-based selection, we used SONAR speech embedding as a query to retrieve examples from the SONAR speech embeddings (sonar\_ss), and from the SONAR text embeddings in the retrieval base (sonar\_st) using nearest neighbors based on the embedding cosine similarity.
 
For audio-based similarity, we utilized embeddings derived from SSL representations for selection. We extracted frame-level audio representations using a pre-trained-only wav2vec-2.0 encoder and then mean-pooled the frame-level representations into a single embedding vector for utterance retrieval (w2v2), employing cosine similarity between embeddings. The embeddings obtained from wav2vec~2.0 representations may be more phonetic than semantic \citep{choi2024self} compared to SONAR embeddings. For text-based similarity, we performed a similarity search based on bm25 \citep{bm25} to select context examples, where we used the target transcript as query (text\_sim) in this case. Note that the text-based similarity baseline cannot be fairly compared to the random selection baseline, as it involves using the target transcript for searching. For selection methods based on similarity to the target, the selected context examples were placed in the order of increasing similarity.

We now turn to the alternative method for constructing context examples based on text in the retrieval base.
In this approach, we selected five examples with the highest unique bigram counts of characters from the retrieval base (bigram), and the same five examples were provided as context examples for all testing audio samples. The bigram selection method maximizes textual diversity within context examples, contrasting with other selection methods that aim to maximize similarity to the target audio. However, the bigram selection method would be biased towards selecting longer context examples, as we did not impose any constraints on the total context length.

For sanity checks and for understanding the capability of the LLM-ASR model, we provided the model with the ``answer,'' setting all five context examples to <target audio><target text> (same\_ex). In this approach, we expect to see significantly improved accuracy compared to all other baselines.

The results averaged on all holdout languages are shown in \Cref{tab:selection-avg}. We consider text\_sim and same\_ex as oracle approaches, as the target transcript is used. Using SONAR embeddings to select examples (sonar\_ss and sonar\_st) yields lower UERs compared to the random selection baseline, reducing CER by up to 11.2\% relative. Using speech-to-speech or speech-to-text embedding retrieval does not show much difference, allowing flexibility to retrieve from either text or speech embeddings.
Using wav2vec~2.0 mean-pooled embeddings for selection does not show obvious improvements over the random baseline. 

The bigram selection yields only a slight improvement over the random baseline, suggesting that the model may struggle to effectively learn from context examples that are not directly related to the target. Moving to the oracle results, having context examples with higher text similarity to the target (text\_sim) shows further gains compared to the SONAR selection baseline. The stronger oracle approach of providing the model with the target audio and transcript pair as context examples (same\_ex) significantly reduces the UER.

From the above results, we can see that even though the model was trained on randomly selected context examples, how we constructed context examples during inference can significantly influence the transcribed text in the zero-shot setting. The oracle results corroborate the fact that the LLM-ASR model can make use of the context examples. From the baseline results, we observe that the model benefits more from examples similar to the target sample over mere textual diversity among context examples.

We present an example of how the transcribed text of the same sample changes with different selection methods in \Cref{tab:example_eng}. 

\begin{table}[!ht]
    \centering
    \begin{tabular}{lccccccc}
    \toprule
    & & & & & & \multicolumn{2}{c}{oracle}  \\
    \cline{7-8}
    & random & sonar\_ss & sonar\_st & w2v2 & bigram & text\_sim & same\_ex \\
    \hline
    MMS-lab & 17.9 & 15.9 & 16.3 & 17.4 & 17.4 & 15.3 & 11.6 \\
    FLEURS & 24.4 & 23.5 & 23.6 & 24.0 & 24.1 & 23.1 & 16.4 \\
    CV & 18.6 & 17.5 & 17.1 & 18.5 & 17.9 & 16.1 & 9.8 \\
    % MLS & & & & & & & \\
    \bottomrule
    \end{tabular}
    \caption{Results for the difference methods of context examples selection. The numbers stand for average UER on the holdout languages.}
    \label{tab:selection-avg}
\end{table}

\begin{table}[!ht]
    \centering
    \begin{tabular}{|l|p{35em}|}
    \hline
        reference text & the school also encourages its students to participate in extracurricular activities via various programmes \\ \hline
        random & the school also encuriges it stoedents to partisipate in ekstra curricular activities wia waries programs \\ \hline
        sonar\_ss & the school also encuriges its students to partisipet in extra curricular activities via veries programs \\ \hline
        same\_ex & the school also encouriges its students to participate in extracuricular activities via various programmes \\
    \hline
    \end{tabular}
    \caption{An example of the transcribed text with different selection methods. English is excluded in the training for this model. Some spelling can be potentially corrected by just changing the context examples provided at inference time. 
    }
    \label{tab:example_eng}
\end{table}

\subsection{Applications to Speech-to-Text Translation}
\label{sec:s2tt_results}

As mentioned at the start of Section 5, we adapted the LLM-ASR variant to perform speech-to-text translation (S2TT) with minimal modification, requiring only the insertion of source and target language identifier (LID) tokens into the input sequence. Despite this simplicity, our experiments show that the model consistently outperforms Whisper and other baselines. Moreover, its performance is comparable to the state-of-the-art SeamlessM4T~\citep{SEAMLESS2025}, which employs a more complex development pipeline specifically designed for speech translation.

\subsubsection{S2TT Experimental Setting}
We first evaluate translation directions of X to English, denoted as X-Eng. For this setting, we used CoVoST2~\citep{wang2020covost} and FLEURS~\citep{conneau2023fleurs} as benchmarks—CoVoST2 covers 21 source languages, while FLEURS spans 101. Our main comparisons are against Whisper and SeamlessM4T~v1.

We reused a large proportion of the X-Eng training data from the SeamlessM4T project.  Following the setup in SeamlessM4T, we do not include FLEURS samples in the training data so that they can serve as a reliable measure of out-of-domain performance. We consider OmniASR-W2V-\{1B, 3B, 7B\} as the encoder when constructing our S2TT models.
Consistent with our LLM-ASR model in \Cref{sec:asr_training_setup}, the decoder is a 1.2B-parameter Transformer in a decoder-only configuration, and we reused the same hyperparameters for training our S2TT models.

\subsubsection{S2TT Results and Discussion}

\begin{table}[htbp]
  \centering
  \begin{tabular}{lrrrr}
    \toprule
    Model  &  \makecell{Model\\Size}  &  \makecell{CoVoST2\\21-Eng}  &  \makecell{FLEURS\\81-Eng}  &  \makecell{FLEURS\\101-Eng} \\
    \midrule
    \textit{Prior Work} \\
    XLSR-2B-S2T~\citep{babu2021xls}    &  2.6B  &  22.1  &    -   &   -     \\
    Whisper Large v2                   &  1.5B  &  29.1  &  17.9  &   -     \\
    SeamlessM4T v1 Medium              &  1.2B  &  29.8  &  20.9  &  18.4   \\
    SeamlessM4T v1 Large               &  2.3B  &  34.1  &  24.0  &  21.4   \\
    AudioPaLM-2-8B-AST~\citep{rubenstein2023audiopalm}   &  8.0B  &  37.8  &  19.7  &   -  \\
    \midrule
    \textit{This Work} \\
    OmniASR-LLM-1B                     &  2.2B  &  34.6  &  19.1  &  16.7   \\
    OmniASR-LLM-3B                     &  4.3B  &  36.7  &  22.1  &  19.4   \\
    OmniASR-LLM-7B                     &  7.7B  &  37.1  &  23.5  &  20.8   \\
    \bottomrule
  \end{tabular}
  \caption{\OmniASR S2TT results in comparison to state-of-the-art speech translation models. We report average BLEU (higher is better) scores across all X-Eng directions on CoVoST2 and FLEURS test splits. Model size indicates the \# of params of that particular model. For Whisper, we started with v3, but its average performance was worse than v2, hence we compared against v2 here.}
  \label{tab:s2tt_results}
\end{table}

Results are presented in \Cref{tab:s2tt_results}, where we also include several baselines in addition to Whisper and SeamlessM4T.  For both CoVoST2 and FLEURS, we report the average BLEU scores across all X-Eng directions on their test sets. Since Whisper only covers~81 out of the~101 to English directions in FLEURS, we also evaluated our models only on these~81 languages to produce a fair comparison against Whisper.

We see that our models largely outperform Whisper on both CoVoST2 and FLEURS, regardless of the model size.
Considering individual language results, we find that our model beats Whisper on~74 out of~81 X-Eng directions on FLEURS.  Compared to SeamlessM4T, our best model outperforms its medium variant across the board, but slightly lags behind its large variant on FLEURS-81 by~0.5 BLEU score point and~0.6 on FLEURS-101. Note that SeamlessM4T initialized its decoder with a pre-trained decoder from NLLB~\citep{NLLB2024}, whereas here we trained our decoder from scratch without any pre-training.the decoder is a 1.2B-parameter Transformer in a decoder-only configuration.

\subsection{Impact of Datamix}
\label{sec:datamix_ablations}

Beyond our primary goal, which is to maximize support for low-resource languages while minimizing regressions in higher-resource ones, we also sought to build robustness against the wide range of noise conditions and speaker variability found in real-world audio. To meet these dual objectives, we designed a series of ablations and upsampling experiments tailored to the challenges of our \allasr{} dataset, which is both highly heterogeneous in audio quality and heavily imbalanced in language coverage.

\subsubsection{Upsampling Low-Resource Languages}

We upsampled at both the corpus- (datasource) and language-levels according to the following hyperparameters: \(\beta_c\)  and \(\beta_l\). \(\beta_c\) determines the relative weight assigned to a particular corpus, and \(\beta_l\) determines the relative weight for a particular language within a corpus. More precisely, for each corpus, we sampled language $L$ according to $p_l \sim \left( \frac{n_l}{N} \right)^{\beta_l}$, where $l = 1,...,L$ is the language, $n_l$ is amount of labeled ASR data for each language within the corpus, and $N$ is total volume of data in the dataset. Sampling across corpora was determined by treating each corpus as a language in the above equation, and using parameter $\beta_c$. This approach is consistent with previous work \citep{pratap2024scaling}. Lower beta values result in higher levels of upsampling of smaller data sources, with 0.0 causing uniform sampling across languages (irrespective of the amount of training data available for each language), and 1.0 representing a baseline where we simply concatenate all data without performing any upsampling.

To determine the optimal upsampling hyperparameters, we performed a sweep across different combinations of $\beta_c$ and $\beta_l$. For hyperparameter selection, we trained a 1B CTC model for 200K steps, and then compared results on all three evaluation protocols described in \Cref{section:eval}: resource-based (\Cref{tab:upsampling_cer_results_resource_bucket}), language-family (\Cref{tab:upsampling_cer_results_lang_family}), and corpus-based (\Cref{tab:upsampling_cer_results_corpus}).

Looking at \Cref{tab:upsampling_cer_results_resource_bucket}, we can see that as we increase language-level upsampling (ie, decrease $\beta_l$ at a given $\beta_c$), CERs decrease for low-resource languages. The baseline (1.0, 1.0) setting, which corresponds to no upsampling, performs by far the worst on low-resource languages. According to results on the resource-based protocol, the best setting is (0.0, 0.0), which is maximal (uniform) upsampling at both the corpus- and language-levels. This setting also gives the highest performance according to the language-grouping evaluation protocol, producing the lowest CERs within each grouping (see \Cref{tab:upsampling_cer_results_lang_family}). 

\Cref{tab:upsampling_cer_results_corpus} shows results on the corpus evaluation protocol. Here, the (0.5, 0.25) setting achieves best results in the corpus evaluation protocol. We can also see here that the (0.0, 0.0) setting obtained lowest CERs on MMS-lab corpus—which comprises over 1000+ languages. This helps explain why it performed so well on the language-based evaluation protocols: they are largely determined by the broad language coverage of MMS-lab. However, this increased MMS-lab performance came at the expense of other datasets such as Babel and CV22, which are known to contain noisier audio data and more diverse speaking conditions. As described subsequently in \Cref{sec:corpus_holdout_ablations}, over-indexing on the narrow audio domain of MMS-lab can have adverse effects on model robustness. Consequently, we chose the (0.5, 0.25) setting when training our final OmniASR models, as this performs well across all corpora and still achieves good results on the language-based protocols.

\begin{table}[htbp]
  \centering
  \begin{tabular}{l r r r r r r | r}
    \toprule
    Condition & Babel & MMS-lab & CV22 & FLEURS\_102 & MLS & OmniASR & Avg \\
    \midrule
    cbeta\_0.0\_lbeta\_0.0   & 27.55 & 4.47 & 17.73 & 9.64 & 3.86 & 24.08 & 14.55 \\
    cbeta\_0.25\_lbeta\_0.5  & 25.05 & 7.07 & 16.74 & 9.24 & 3.25 & 25.75 & 14.52 \\
    cbeta\_0.5\_lbeta\_0.5   & 25.71 & 6.32 & 17.14 & 9.63 & 3.26 & 26.23 & 14.71 \\
    cbeta\_0.75\_lbeta\_0.5  & 27.01 & 5.82 & 17.10 & 10.46 & 3.32 & 27.54 & 15.21 \\
    cbeta\_0.5\_lbeta\_0.25  & 25.41 & 6.05 & 16.42 & 9.42 & 3.32 & 26.08 & 14.45 \\
    cbeta\_0.5\_lbeta\_0.75  & 25.85 & 6.55 & 17.94 & 9.61 & 3.20 & 26.35 & 14.92  \\
    cbeta\_1.0\_lbeta\_1.0   & 28.72 & 6.09 & 21.30 & 11.15 & 3.20 & 29.33 & 16.63 \\
    \bottomrule
  \end{tabular}
  \caption{Performance (CER) across dev splits for each corpus in AllASR dataset for different $beta$ values. The rightmost column (avg) is separated for clarity.}
  \label{tab:upsampling_cer_results_corpus}
\end{table}

\begin{table}[htbp]
  \centering
  \begin{adjustbox}{max width=\textwidth}
  \begin{tabular}{l r r r r r r r}
    \toprule
    \multicolumn{8}{c}{\textbf{CERs for (cbeta, lbeta) upsampling}} \\
    \midrule
    \makecell{Language\\Groupings} & (0.0, 0.0) & (0.25, 0.5) & (0.5, 0.25) & (0.5, 0.5) & (0.5, 0.75) & (0.75, 0.5) & (1.0, 1.0) \\
    \midrule
    Afroasia   & 16.35 $\pm$ 3.11 & 18.70 $\pm$ 3.20 & 18.06 $\pm$ 3.18 & 18.45 $\pm$ 3.40 & 18.86 $\pm$ 3.94 & 17.96 $\pm$ 3.20 & 19.37 $\pm$ 4.12 \\
    Amazbasi   & 3.12 $\pm$ 0.41  & 5.04 $\pm$ 0.58  & 4.34 $\pm$ 0.52  & 4.39 $\pm$ 0.52  & 4.48 $\pm$ 0.55  & 4.13 $\pm$ 0.51  & 4.12 $\pm$ 0.57  \\
    Amerande   & 3.40 $\pm$ 0.72  & 4.95 $\pm$ 0.85  & 4.45 $\pm$ 0.85  & 4.55 $\pm$ 0.87  & 4.63 $\pm$ 0.88  & 4.48 $\pm$ 0.96  & 4.65 $\pm$ 1.13  \\
    Atlacong   & 12.43 $\pm$ 1.19 & 15.47 $\pm$ 1.15 & 14.70 $\pm$ 1.19 & 14.90 $\pm$ 1.19 & 15.05 $\pm$ 1.19 & 14.95 $\pm$ 1.25 & 15.62 $\pm$ 1.34 \\
    Austasia   & 11.80 $\pm$ 6.55 & 14.41 $\pm$ 6.32 & 12.88 $\pm$ 7.14 & 13.52 $\pm$ 7.14 & 13.98 $\pm$ 7.37 & 13.43 $\pm$ 7.29 & 14.19 $\pm$ 6.90 \\
    Austrone   & 6.63 $\pm$ 1.10  & 8.07 $\pm$ 1.11  & 7.76 $\pm$ 1.14  & 7.88 $\pm$ 1.14  & 7.99 $\pm$ 1.15  & 7.99 $\pm$ 1.20  & 8.35 $\pm$ 1.25  \\
    Caucasus   & 11.89 $\pm$ 4.39 & 11.95 $\pm$ 3.46 & 13.62 $\pm$ 5.10 & 12.58 $\pm$ 4.33 & 13.76 $\pm$ 5.24 & 13.13 $\pm$ 5.39 & 14.43 $\pm$ 5.16 \\
    Dravidia   & 13.16 $\pm$ 8.77 & 14.26 $\pm$ 7.57 & 13.73 $\pm$ 7.89 & 14.02 $\pm$ 7.73 & 14.02 $\pm$ 7.15 & 13.99 $\pm$ 7.80 & 14.08 $\pm$ 6.70 \\
    Indoeuro   & 13.15 $\pm$ 1.95 & 14.47 $\pm$ 1.99 & 14.35 $\pm$ 2.00 & 14.74 $\pm$ 2.02 & 15.15 $\pm$ 2.15 & 15.15 $\pm$ 2.09 & 17.25 $\pm$ 2.28 \\
    Mesoamer   & 10.53 $\pm$ 1.93 & 13.05 $\pm$ 1.88 & 12.36 $\pm$ 1.93 & 12.55 $\pm$ 1.92 & 12.67 $\pm$ 1.91 & 12.59 $\pm$ 1.99 & 13.21 $\pm$ 2.09 \\
    Newguine   & 7.27 $\pm$ 2.31  & 9.24 $\pm$ 2.39  & 8.68 $\pm$ 2.44  & 8.83 $\pm$ 2.44  & 8.97 $\pm$ 2.45  & 8.76 $\pm$ 2.55  & 9.07 $\pm$ 2.65  \\
    Nilosaha   & 7.23 $\pm$ 1.81  & 10.36 $\pm$ 1.76 & 9.25 $\pm$ 1.85  & 9.46 $\pm$ 1.86  & 9.69 $\pm$ 1.89  & 9.25 $\pm$ 2.01  & 9.51 $\pm$ 2.17  \\
    Norameri   & 8.22 $\pm$ 3.88  & 11.32 $\pm$ 3.77 & 10.08 $\pm$ 4.03 & 11.16 $\pm$ 4.40 & 12.11 $\pm$ 6.07 & 10.55 $\pm$ 4.24 & 13.40 $\pm$ 7.81 \\
    Sinotibe   & 13.72 $\pm$ 4.91 & 15.85 $\pm$ 4.93 & 14.88 $\pm$ 4.97 & 15.22 $\pm$ 5.03 & 15.85 $\pm$ 5.30 & 15.50 $\pm$ 5.34 & 16.97 $\pm$ 5.90 \\
    Misc       & 23.05 $\pm$ 2.46 & 24.54 $\pm$ 2.53 & 24.56 $\pm$ 2.54 & 24.82 $\pm$ 2.52 & 25.13 $\pm$ 2.50 & 25.68 $\pm$ 2.55 & 27.47 $\pm$ 2.57 \\
    \midrule
    \textbf{Average} & 10.80 $\pm$ 3.03 & 12.78 $\pm$ 2.90 & 12.25 $\pm$ 3.12 & 12.47 $\pm$ 3.10 & 12.82 $\pm$ 3.32 & 12.50 $\pm$ 3.22 & 13.44 $\pm$ 3.51 \\
    \bottomrule
  \end{tabular}
  \end{adjustbox}
  \caption{Performance (CER) across language groupings for different upsampling conditions. CER is averaged across all languages within each language family; error bars indicate 95\% Confidence Intervals.}
  \label{tab:upsampling_cer_results_lang_family}
\end{table}

\begin{table}[htbp]
  \centering
  \begin{tabular}{l r r r | r}
    \toprule
    Condition & High & Med & Low & \textbf{Avg} \\
    \midrule
    cbeta\_0.0\_lbeta\_0.0   & 6.28 & 6.12 & 21.14 & 11.18 \\
    cbeta\_0.25\_lbeta\_0.5  & 6.80 & 8.70 & 23.23 & 12.91 \\
    cbeta\_0.5\_lbeta\_0.5   & 6.54 & 8.08 & 23.42 & 12.5 \\
    cbeta\_0.75\_lbeta\_0.5  & 6.60 & 7.63 & 24.38 & 12.68 \\
    cbeta\_0.5\_lbeta\_0.25  & 6.54 & 7.86 & 23.09 & 12.89 \\
    cbeta\_0.5\_lbeta\_0.75  & 6.50	& 8.37 & 23.79 & 12.87 \\
    cbeta\_1.0\_lbeta\_1.0   & 6.75 & 8.09 & 26.40 & 13.75 \\
    \bottomrule
  \end{tabular}
  \caption{Performance (CER) across resource buckets and conditions for different $beta$ values. The rightmost column (Avg) is separated for clarity.}
  \label{tab:upsampling_cer_results_resource_bucket}
\end{table} 

\subsubsection{Generalizing to Unseen Audio Distributions}
\label{sec:corpus_holdout_ablations}

In addition to optimizing for low-resource languages, we also wanted to ensure our model was robust to various audio conditions. As such, we ran an ablation where we trained models on the \allasr{} dataset, holding out one corpus at a time. Here $AllASR$ refers to: MMS-lab, \vendorcorpus, OMSF, FLEURS-102, Babel, MLS, and CV22\footnote{Note this is a subset of \allasr{} used to train our final models. Refer to Table~\ref{tab:all_asr_dataset} for a list of all data sources}. For example, $AllASR\_x\_mls$ refers to a model trained on all of the above except MLS. We evaluates these $AllASR\_x$ holdout models on development splits from the held-out data sources, and compares them to a baseline model trained on the complete $AllASR$ dataset, thus measuring their ability to generalize to unseen audio distributions. 

Further, we contrasted these hold-out model conditions with a model trained on just MMS-lab. This latter model was still exposed to all languages in the hold-out sources, but it was not exposed audio from any other data sources. Comparing the $AllASR\_x$ holdout models against MMS-lab model allows us to assess the degree to which our model becomes better at generalizing to new audio distributions as we expand the training set to include more sources. In all conditions, we trained 1B CTC models for 100K steps at 32 GPUs.

\begin{table}
    \centering
    \begin{tabular}{llrrr}
    \toprule
         Training Data&  Holdout Data source&  Holdout CER&  Baseline CER& Baseline CERR\\
    \midrule
         AllASR\_x\_mls&  mls&  4.3&  3.27& -31\%\\
         AllASR\_x\_fleurs&  fleurs&  20.95&  11.95& -75\%\\
         AllASR\_x\_cv22&  cv22&  33.46&  19.57& -71\%\\
         MMS-lab&  mls&  6.34&  3.27& -94\%\\
         MMS-lab&  fleurs&  35.72&  11.95& -199\% \\
 MMS-lab& cv22& 43.35& 19.57&-1.22\\
    \bottomrule
    \end{tabular}
    \caption{Corpus holdout ablation results. Rows 1-3 contain performance for models  trained on our AllASR dataset, with a single source held-out from training. CERs for heldout corpora are shown in third column, and can be compared to CER obtained by a baseline model trained on all data (including the holdout corpus, fourth column). Rows 4-6 show holdout performance of a model trained on just MMS-lab, which covered all languages in the holdout corpora but had less audio diversity. Column 5 shows relative Character Error Rate reduction (CERR) of the holdout condition relative to the baseline: $(CER_{baseline} - CER_{treatment})/CER_{baseline}$. These values are all negative, indicating regressions for the holdout models compared to the baseline, which has seen all the data.}
    \label{tab:corpus_holdout_ablation_results}
\end{table}

Results are displayed in \Cref{tab:corpus_holdout_ablation_results}. Rows 1-3 show CERs obtained by $AllASR\_x$ models on their respective holdout corpora (column 3). These numbers can be compared against Baseline CERs obtained by the $AllASR$ model (column 4). Baseline CERR (column 5) makes this delta explicit: more negative values indicate larger regressions compared to baseline. As expected, performance regresses for all held-out data sources compared to the baseline.  The regression is more pronounced on FLEURS and CV22 than on MLS, suggesting that those two sources comprise more distinct audio distributions compared to the other sources within AllASR. That said, models still perform reasonably well on the holdout corpora (especially on MLS), indicating an ability to generalize to unseen audio distributions.

Crucially, Baseline CERR is substantially better in the $AllASR\_x$ models compared to the MMS-lab condition. This is true across all three holdout sources and indicates that our $AllASR$ recipe improves our model's ability to generalize to unseen audio distributions, as compared to training on a single data source with the same language coverage.

\subsubsection{Model Robustness to Background Noise}
\label{sec:noise_robustness}

\begin{figure}
    \centering
    \includegraphics[width=0.75\linewidth]{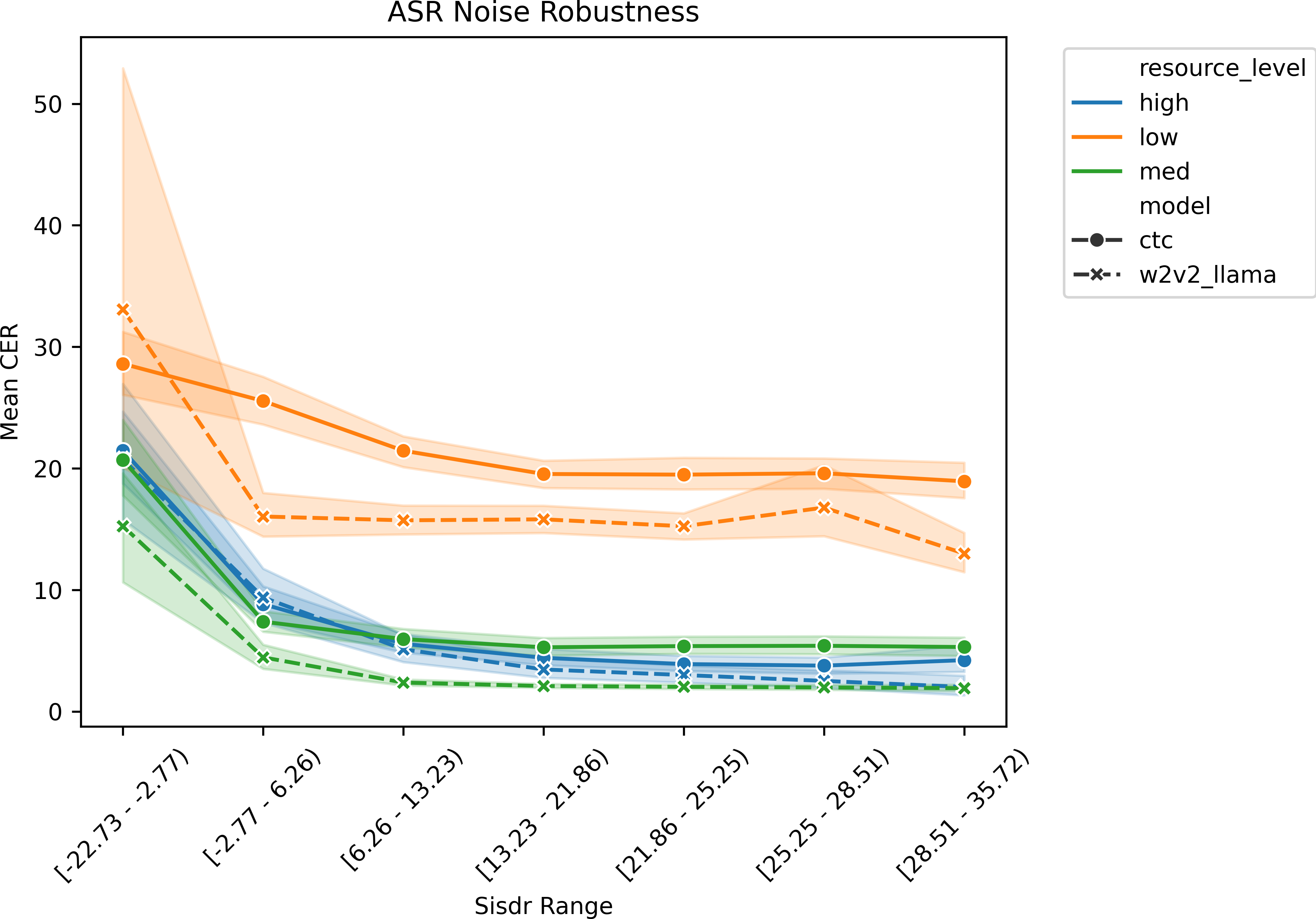}
    \caption{ASR noise robustness across All\_ASR dev sets. Utterances were binned into SI-SDR ranges that showcase the outliers with low SI-SDR values, up through the rest of the distribution. Mean CER values, averaged across languages (y-axis), are plotted against SI-SDR range (x-axis) of the associated audio. Error ribbons indicate 95\% CI. Results are further grouped by resource-level of the included language, as indicated by color: low- (orange; <10 hrs), medium- (blue; 10<=hours<50 hours), and high-resource (green; >50 hours). Results presented are for the 7B CTC model (solid line) and w2v2\_LLM (dashed line). Error ribbons indicate 95\% Confidence Intervals.}
    \label{fig:cer_vs_sisdr}
\end{figure}

Building on the previous section, we further examine model robustness by measuring ASR performance as a function of background noise and/or clarity of the speech signals. To do this, we ran audio samples in our development sets through the Torchaudio Squim models, which emit estimations of speech audio quality \citep{kumar2023squim}. \Cref{fig:cer_vs_sisdr} shows CER as a function of SI-SDR, which is a model estimate of the level of background noise relative to the speech signal. Model performance on different language groups is shown in different colors according to the resource-level (number of training hours) associated with each language. The analysis was performed on our 7B CTC model (solid line) as well as our 7B LLM-ASR model (dashed line).

Results are presented in \Cref{fig:cer_vs_sisdr}. Each utterance was binned into SI-SDR ranges, which were not evenly spaced but instead selected to showcase the extreme outliers in the distribution of our \allasr{} dataset (i.e., audios with large amounts of background noise). The ranges correspond to the following SI-SDR percentiles: [0-1, 1-5, 5-20, 20-40, 40-60, 60-80, 80-95, 95-100].  To remove any confounds with LID, we only include languages with utterances in each of the displayed SI-SDR bins. Within each SI-SDR bin, we obtain Mean CER (averaged across languages; y-axis) and plot this against SI-SDR bin-range (x-axis). Error ribbons indicate 95\% Confidence Intervals. CTC performance is indicated by dots/solid line, while LLM-ASR is indicated by x/dashed line. Languages with different levels of training data are grouped by color: low-resource (<10 hours), medium-resource (between 10-50 hours), and high-resource (>50 hours). 

As expected, CER is higher for utterances with low SI-SDR values (high background noise) compared to utterances with higher SI-SDR (cleaner audio). CER is highest and most variable at the extreme low-end (lowest 1\% of SI-SDR).  However, CER quickly drops and flattens out after this. For instance, even for the noisiest 1\%-5\% of utterances, LLM-ASR model obtains CERs $\leq 10$ across all language groups, and the CTC model obtains CERS $<15$ for medium- and high-resource languages. In the remaining SI-SDR bins, CER is quite flat within each language group. It is important to recall that the x-axis in \Cref{fig:cer_vs_sisdr} is not a linear scale throughout: the first two bin-ranges represent outlier utterances with extreme levels of background noise (i.e., top 1\% and top 5\%, respectively). Overall, these results indicate good model robustness to moderate levels of background noise (i.e., lowest 5\% percentiles), and that our models do not exhibit any bias in background noise sensitivity as a function of language resource-level.

\subsubsection{\vendoromsf Holdout Ablation}
\label{sec:omni_asr_corpus_holdout}

To measure the value of the \vendoromsf data (i.e., all the new data collected in this project: \vendorcorpus plus OMSF), we ran a simple ablation in which we compared a \allasr{} model against an \allasr{}\_x\_\vendoromsf model.  In the latter, we held out \vendoromsf data from training and then evaluated the model on the hold-out \vendoromsf dev sets. In both conditions, we trained 7B LLM-ASR models for 150K steps across 64 GPUs.

To be clear, \vendoromsf introduces mostly new languages to the mix, so in these cases, the \allasr{}\_x\_\vendoromsf is being evaluated on languages it was not exposed to during training. In these cases, we expect the \allasr{} model to outperform the holdout model. Nevertheless, we include the ablation results to validate the training signal in \vendoromsf; this allows us to ensure that our claims of supporting newly introduced languages are well founded.

Additionally, there are 13 overlapping languages in \vendoromsf that are also contained in other corpora within \allasr{}. For these languages, we would like to see if the additional \vendoromsf training data provides a valuable signal above and beyond what was already present in our training data, especially with regard to speaker diversity and more naturalistic audio conditions. We separately report ablation results for new and overlapping languages in \Cref{tab:omni_holdout_ablation}.

\begin{table}
    \centering
    \begin{tabular}{rll}
        \toprule
         Data Condition & \makecell{CER\\(new languages)} & \makecell{CER\\(overlap langs)} \\
         \midrule
         $AllASR\_x\_OMNI$ & 47.03 & 39.46 \\
         $AllASR\_OMNI$ & 22.62 & 11.50 \\
    \bottomrule
    \end{tabular}
    \caption{\vendoromsf holdout ablation results. Mean CERs on the \vendoromsf dev sets, averaged across languages are shown for the holdout and full-data conditions. Results reported separately for new languages introduced by \vendoromsf versus languages that were already present in other corpora within \allasr{}.}
    \label{tab:omni_holdout_ablation}
\end{table}

Results in \Cref{tab:omni_holdout_ablation} highlight the value of the \vendoromsf data collected in this project, both by extending coverage to new languages and by substantially improving performance on already-supported ones. For new languages, our AllASR\_OMNI model achieves a mean CER of 22.62, less than half the 47.03 obtained by the holdout model. Although 22.62 remains relatively high compared to CERs obtained on other corpora, it nevertheless represents a major reduction from the holdout model’s zero-shot performance, despite that model being highly multilingual. For overlapping languages, the impact of \vendoromsf data is even more striking: CERs drop from 39.46 with the holdout model to 11.50 with AllASR\_OMNI.

This latter result underscores the fact that data from \vendoromsf is quite challenging for ASR compared to many pre-existing multilingual datasets, which mostly consist of clean, studio-quality recordings of speaker-reading. \vendoromsf was intentionally curated to represent naturalistic (i.e., often noisy) audio conditions, diverse speaker identities, and spontaneous, expressive speech. The benefits of such data are demonstrated here: without including them in the datamix, an equally multilingual model (i.e., our holdout) struggles in these more difficult, but more naturalistic audio/speaker conditions. In sum, by including \vendoromsf, we introduce new language coverage and also substantially improve model robustness, which ultimately situates our models for use in the wild.

\subsubsection{Fine-tuning for Individual Low-Resource Languages}
\label{sec:language_specific_ablations}

In this study, we fine-tuned bespoke CTC models on individual low-resource languages. There are two motivations here. First, from a theoretical standpoint, we are interested in establishing the best performance achievable for languages with fewer than 10 hours of data, and in quantifying the performance gap relative to our \OmniASR{} models trained across 1,600+ languages. Second, we present our learnings to the community to provide recommended settings for users interested in adapting and optimizing our open-source models for their own bespoke purposes, especially in lower compute settings. This study was performed with 11 low-resource languages, with between 5-10 hours of training data and at least 1 hour of validation splits. See \Cref{tab:low_res_lang_list} for the complete list.

\begin{table}
    \centering
    \begin{tabular}{llrr}
    \toprule
         LID & Script &  \# train hours& Best CER\\
    \midrule
         ast&Latn& 8.1 & 3.31 \\
         ckb&Arab& 9.1 & 4.47 \\
         ltz&Latn& 8.5 & 7.42 \\
         hsb&Latn& 9.1 & 2.17 \\
         afo&Latn& 7.6 & 29.7 \\
         ahl&Latn& 7.2 & 16.47 \\
         div&Thaa& 7.6 & 5.16 \\
         fuv&Latn& 6.5 & 15.1 \\
         qxp&Latn& 9.9 & 1.61 \\
         ajg&Latn& 9.5 & 8.05 \\
         vro&Latn& 9.5 & 5.7 \\
    \bottomrule
    \end{tabular}
    \caption{Low-resource languages used in language-specific study.}
    \label{tab:low_res_lang_list}
\end{table}

We fine-tuned language-specific CTC models for each of these 11 languages, across the 300M, 1B, and 3B scales. In one condition, we seeded from a pretrained w2v2 checkpoint, and in another, we seeded from an OmniASR CTC checkpoint, which was pretrained on all 1600+ languages. For the w2v2-seed condition, we trained with a learning rate of 1e-05 for 30K steps, though we observed that models typically converge within 10K steps. For the CTC-seed condition, we also use an lr of 1e-05 and trained for 5K steps. CTC fine-tuning takes \~1 hour of walltime on 32 GPUs for the 300M scale. These hyperparameters were selected based on empirical sweeps for a couple of exemplar languages, but of course, in practice, the optimal training hyperparameters will be a function of the specific language and data used in finetuning. For example, we observed that certain languages converged long before the \# training steps listed here.

We then compare the performance (CER) of these language-specific models against our \OmniASR CTC models at each scale. These \OmniASR models were trained on all 1600+ languages, without any sort of language-specific optimization. Results can be found in \Cref{tab:cer_per_low_res_language}. Language-specific models substantially outperform the \OmniASR baselines, achieving CERs of less than 5 in many of these low-resource languages—even at the smaller 300M and 1B scales. Additionally, CTC-seeded models consistently outperformed w2v2-seeded models at the 300M and 1B scales, even though they were fine-tuned for a fraction of the training steps (5K instead of 30K). Consequently, we advise practitioners wishing to optimize our 300M and 1B models for ASR in particular low-resource languages to seed with CTC checkpoints. However, at the 3B scale the w2v2-seeded checkpoints trained for 30K steps generally outperformed the ctc-seeded checkpoints trained for 5K steps.

\Cref{tab:cer_per_low_res_language} also shows CERs obtained by our 7B OmniASR LLM model in the rightmost column. In most cases, the OmniASR 7B-LLM was quite competitive with the language-specific models, indicating an extremely high performance on these low-resource languages despite the fact that it was trained on all 1600+ languages and without any language-specific optimization. On the other hand, even though the language-specific models are significantly smaller than the 7B-LLM model and lack the LLM architectural component, they still obtained lower CERs for most languages, even at the smallest 300M scale. This demonstrates a unique strength of our open-source \OmniASR models: they contain rich omnilingual knowledge, and can be quickly adapted and fine-tuned to excel in particular low-resource settings with minimal compute. Once fine-tuned, the lightweight CTC models can be run in small compute environments during inference, which can be desirable in numerous applications.

\begin{table}[htbp]
  \centering
  \begin{tabular}{ll
                  >{\raggedleft\arraybackslash}p{1.7cm}
                  >{\raggedleft\arraybackslash}p{1.7cm}
                  r | r}
    \toprule
    Language & Scale & \multicolumn{2}{c}{Single-Lang} & OmniCTC & \multicolumn{1}{c}{OmniLLM (7B)} \\
    \cmidrule(lr){3-4}
    & & CTC Seed & W2V2 Seed & & \\
    \midrule
    afo\_Latn & 300m & 32.54 & 32.32 & 33.54 & 38.91 \\
              & 1b   & 31.58 & 29.71 & 33.17 &       \\
              & 3b   & 30.89 & 29.11 & 32.18 &       \\
    \cmidrule(lr){1-6}
    ahl\_Latn & 300m & 18.78 & 20.52 & 44.28 & 24.33 \\
              & 1b   & 17.66 & 16.47 & 36.76 &       \\
              & 3b   & 17.87 & 15.27 & 34.61 &       \\
    \cmidrule(lr){1-6}
    ajg\_Latn & 300m & 8.05  & 8.63  & 21.97 & 7.54  \\
              & 1b   & 8.82  & 8.11  & 19.14 &       \\
              & 3b   & 9.02  & 7.92  & 15.63 &       \\
    \cmidrule(lr){1-6}
    ast\_Latn & 300m & 4.95  & 8.02  & 10.87 & 5.105 \\
              & 1b   & 3.55  & 4.83  & 7.88  &       \\
              & 3b   & 3.91  & 3.31  & 6.44  &       \\
    \cmidrule(lr){1-6}
    ckb\_Arab & 300m & 5.82  & 8.01  & 15.29 & 4.73  \\
              & 1b   & 5.05  & 5.91  & 12.28 &       \\
              & 3b   & 5.20  & 4.17  & 9.94  &       \\
    \cmidrule(lr){1-6}
    div\_Thaa & 300m & 5.54  & 8.36  & 19.21 & 5.58  \\
              & 1b   & 5.16  & 5.66  & 17.21 &       \\
              & 3b   & 5.45  & 4.57  & 13.04 &       \\
    \cmidrule(lr){1-6}
    fuv\_Latn & 300m & 16.41 & 18.45 & 23.69 & 26.83 \\
              & 1b   & 15.59 & 15.10 & 20.47 &       \\
              & 3b   & 15.14 & 14.35 & 16.31 &       \\
    \cmidrule(lr){1-6}
    hsb\_Latn & 300m & 2.93  & 7.18  & 10.41 & 4.1   \\
              & 1b   & 2.57  & 2.17  & 7.07  &       \\
              & 3b   & 3.20  & 1.79  & 4.94  &       \\
    \cmidrule(lr){1-6}
    ltz\_Latn & 300m & 9.88  & 15.94 & 19.72 & 6.07  \\
              & 1b   & 7.42  & 10.72 & 12.44 &       \\
              & 3b   & 8.09  & 7.12  & 9.80  &       \\
    \cmidrule(lr){1-6}
    qxp\_Latn & 300m & 1.70  & 2.08  & 4.49  & 1.32  \\
              & 1b   & 1.61  & 1.68  & 2.94  &       \\
              & 3b   & 1.81  & 1.47  & 2.71  &       \\
    \cmidrule(lr){1-6}
    vro\_Latn & 300m & 7.18  & 9.39  & 16.74 & 4.02  \\
              & 1b   & 6.36  & 5.70  & 12.67 &       \\
              & 3b   & 6.76  & 5.12  & 10.16 &       \\
    \bottomrule
  \end{tabular}
  \caption{Model performance (CER) across low-resource languages and scales. Columns 3-4 show language-specific models. The rightmost column (OmniLLM (7B)) is separated for clarity.}
  \label{tab:cer_per_low_res_language}
\end{table}

\subsection{Impact of Conditioning on Language Codes}   
\label{sec:lid_results}

\begin{table}[ht]
\centering
\begin{tabular}{llrrrrrrrr}
\toprule
Language  & Conditioning & MMS-Lab & Omnilingual ASR & Babel & FLEURS-102 & MLS & CV22 \\
Conditioning & at Inference \\
\midrule
       0.0 & No    & 2.5  &	13.3 &	19.1 &	7.9 &	2.6 &	11.3 \\
       0.2 & No    & 2.5	& 13.4	& 19.3	& 7.4	& 2.6	& 11.8 \\
       & Yes               & 2.5 & 	13.2	& 19.2	& 7.6	& 2.6& 	8.2 \\
       0.5 & No            & 2.5 &	13.7	& 19.4 &	7.5 &	2.6 &	11.8 \\
       & Yes               & 2.5	& 13.4	& 19.3	& 7.1 &	2.6 &	7.9 \\
       1.0 & No            & 15.7	& 42.5	& 54.1	& 34.7	& 3.1	& 45.1 \\
       & Yes               & 2.5	& 14.0	& 19.2	& 6.9	& 2.6	& 6.9 \\
\bottomrule
\end{tabular}
\caption{Impact of language and script conditioning on the LLM-ASR model. A model with language and script conditioning 50\% of the time during training is able to deliver best tradeoff between inference modes—when language and script information are either absent or provided.}
\label{tab:lid}
\end{table}

We performed an ablation experiment to study the impact of conditioning the model on the ID of the language and script combination as described in \Cref{sec:lid}. Models trained with this feature can be evaluated with or without providing the language and script information. To measure its effect, we compared a model trained without language and script ID conditioning against models trained with different probabilities of including this information during training. 

The results in \Cref{tab:lid} show that compared to a baseline trained without conditioning, training with language and script conditioning on at least 50\% of the samples yields considerable improvements on FLEURS-102 and Common Voice when conditioning is used at inference. These accuracy gains largely come from utterances that, without conditioning, were misrecognized in the wrong language or script—errors that significantly increased CER. Importantly, training with conditioning applied to only half of the batches preserved the model’s ability to operate effectively without conditioning at inference, still recognizing the correct language and script for the vast majority of samples. In fact, this setup showed virtually no degradation in accuracy compared to the baseline model (training language conditioning for 0\% of the samples) when conditioning was not applied at inference. Based on these findings, we adopt language and script conditioning for 50\% of the samples during training in our final LLM-ASR models.

\subsection{Comparison of OmniASR-W2V Models to Existing SSL Speech Encoders}
\label{sec:ssl_compare}

In this section, we compare the OmniASR-W2V family with some of the most widely used multilingual SSL speech encoders, including XLSR-\{0.3B, 1B, 2B\} from~\citet{babu2021xls} and MMS-\{0.3B, 1B\} from~\citet{pratap2024scaling}.  In \Cref{tab:ssl_config_compare}, we highlight the key differences among the models, focusing on the number of languages covered, the volume of pre-training data, and the model size measured in parameters.

\begin{table}[htbp]
  \centering
  \begin{tabular}{lrlrr}
    \toprule
    Model & \# of lang & Datasets & \makecell{Data\\volume~(hrs)} & \# of params \\
    \midrule
    \textit{Prior Work} \\
    XLSR-0.3B   &   128  &  VP, MLS, CV6, VL, BBL  &  436k  &   317M \\
    XLSR-1B     &   128  &  VP, MLS, CV6, VL, BBL  &  436k  &   965M \\
    XLSR-2B     &   128  &  VP, MLS, CV6, VL, BBL  &  436k  &  2162M \\
    MMS-0.3B    &  1406  &  VP, MLS, CV9, VL, BBL, MMS-Lab, FL  &  491k  &  317M \\
    MMS-1B      &  1406  &  VP, MLS, CV9, VL, BBL, MMS-Lab, FL  &  491k  &  965M \\
    \midrule
    \textit{This Work}  \\
    OmniASR-W2V-0.3B   &  1600+  &  \ssldata~(Section~\ref{sec:ssl_data}) &  4.3M  &  317M \\
    OmniASR-W2V-1B     &  1600+  &  \ssldata  &  4.3M  &   965M \\
    OmniASR-W2V-3B     &  1600+  &  \ssldata  &  4.3M  &  3046M \\
    OmniASR-W2V-7B     &  1600+  &  \ssldata  &  4.3M  &  6488M \\
    \bottomrule
  \end{tabular}
  \caption{Existing SSL speech encoders. VP, MLS, CV, VL, BBL, and FL stand for VoxPopuli, Multilingual LibriSpeech, Common Voice, VoxLingua, Babel, and FLEURS, respectively.  Note that XLSR and MMS models used different versions of CV: CV6 and CV9, where the latter covers~29 more languages.}
  \label{tab:ssl_config_compare}
\end{table}

To enable a fair comparison, all pre-trained speech encoders were fine-tuned with CTC on \allasr following the setting specified in~\Cref{sec:asr_training_setup}.  We report the test set results on MMS-Lab, \vendorcorpus, FLEURS-102, MLS, and CV22 in~\Cref{tab:ssl_result_compare}.

\begin{table}[htbp]
  \centering
  \begin{tabular}{lccccc}
    \toprule
    Model  &  MLS  &  FLEURS-102  &  MMS-Lab  &  CV22  &  \vendorcorpus \\
    \midrule
    \textit{Prior Work} \\
    XLSR-0.3B         &  3.7  &  14.6  &  12.6  &  24.0  &  30.3 \\
    XLSR-1B           &  2.9  &  10.2  &   7.6  &  18.8  &  25.7 \\
    XLSR-2B           &  3.0  &   9.9  &   5.8  &  19.5  &  24.5 \\
    MMS-0.3B          &  4.1  &  14.2  &   8.2  &  22.2  &  29.1 \\
    MMS-1B            &  3.2  &  10.2  &   4.7  &  16.8  &  25.2 \\
    \midrule
    \textit{This Work}  \\
    OmniASR-W2V-0.3B  &  4.1  &  12.0  &   7.3  &  20.2  &  26.4 \\
    OmniASR-W2V-1B    &  3.1  &   8.9  &   4.5  &  16.5  &  24.1 \\
    OmniASR-W2V-3B    &  2.7  &   8.0  &   3.5  &  16.2  &  22.8 \\
    OmniASR-W2V-7B    &  2.5  &   7.5  &   3.1  &  15.8  &  20.8 \\
    \bottomrule
  \end{tabular}
  \caption{Results of existing SSL speech encoders and the OmniASR-W2V models. For each benchmark, we report the average CER across languages on the test set.}
  \label{tab:ssl_result_compare}
\end{table}

Comparing models of the same size, we see that OmniASR-W2V-0.3B outperforms XLSR-0.3B and MMS-0.3B on all benchmarks except for MLS, where OmniASR-W2V-0.3B's performance is on par with MMS-0.3B but worse than XLSR-0.3B.  Note that while XLSR-0.3B outperforms OmniASR-W2V-0.3B by less than~10\% on MLS, its performance on the rest of the benchmarks lags behind OminASR-W2V-0.3B by~18\%, 42\%, 16\%, and 13\%, respectively.  A similar conclusion can be drawn from the comparison of OmniASR-W2V-1B, XLSR-1B, and MMS-1B, except for the fact that, now, OmniASR-W2V-1B beats MMS-1B in all cases, and the performance gap with XLSR-1B on MLS is reduced to~6\%.

Scaling beyond~1B, we see OmniASR-W2V-3B and OmniASR-W2V-7B continue to widen the gap with other encoders across all benchmarks, suggesting they are the best choices for optimal performance on both top languages and long-tailed languages.

\section{Societal Impact and Conclusion}
\label{section:impact}

\OmniASR illustrates how scaling methods, when combined with deliberate data collection and new architectural innovation, can reshape the trajectory of multilingual ASR. The project not only extends coverage to more than 1,600 languages, with over 500 represented for the first time in any ASR system, but also reframes how coverage itself is conceived. In contrast to prominent existing systems \citep{radford2023robust,pratap2024scaling,zhang2023google}, where unsupported languages could only be added through expert-driven fine-tuning, \OmniASR demonstrates that recognition can be extended to entirely new languages with just a few in-context samples. This shift from fixed coverage to open-ended extensibility enables certain underserved groups to bring their languages into conversation with digital tools that have historically excluded them.

The coexistence of massive, high-accuracy models with lightweight 300M-parameter variants also alters the economics of deployment, making it feasible to adapt ASR both to high-compute cloud infrastructures and to low-power devices in areas with limited connectivity. This flexibility broadens not only the range of research questions that can be pursued but also the contexts in which ASR can be applied, from speech-to-text translation pipelines to community-led archives. By open-sourcing models and training pipelines, \OmniASR lowers the barriers to entry, shifting long-tail ASR research from a niche pursuit to a tractable and collaborative enterprise.

For language communities, the impact is both promising and contingent. Already, \OmniASR is being deployed in practice: health practitioners in Nigeria are using the system to facilitate Hausa transcriptions in community clinics, with the intention of improving documentation and patient care. In oral cultures, it could help make endangered archives more searchable; in education, lightweight models might power interactive learning tools in mother tongues; in civic life, transcription of local-language broadcasts could expand access to news and information. Yet these same capabilities can also be repurposed in ways that conflict with community priorities, from surveillance to unwanted moderation \citep{abdullah2021sok}. This tension underscores the need for participatory governance and ongoing dialogue, rather than one-time transfers of technology \citep{wang2024human}.

Importantly, our community partners remind us of the need for large technology companies not only to draw on open language data but also to reinvest in its creation and stewardship. \OmniASR was designed in this spirit: not as an act of charity, but as part of a healthy, respectful, and mutually beneficial ecosystem in which communities are compensated for the time and emotional labor that language documentation entails. In light of ongoing discussions about consent and compensation in AI training data, it is essential to acknowledge that these concerns highlight the complexities surrounding ethical practices in this field of research. They point to longstanding issues of power, participation, and equity in how language resources are built and shared. Our approach—compensating native speakers and working through local partnerships—was one attempt to respond to these challenges. Still, compensation should not be seen as a panacea: some communities may prefer voluntary, crowdsourced participation, while others may feel financially pressured into contributing data. Although we did not observe such dynamics in our own experience, they remain a possibility and highlight the importance of vigilance in future work to ensure that participation is informed, voluntary, and aligned with community priorities.

Reflecting on the project’s trajectory, several broader lessons emerge. First, the long tail of languages should not be treated as a final frontier to be “solved” once and for all, but as a dynamic, evolving space of collaboration in which linguistic, technical, and social knowledge interact. Second, open-sourcing at this scale is not merely an act of transparency but an intervention that redistributes the power to innovate, enabling actors historically excluded from large-scale AI development. Third, large-scale ASR is inseparable from the politics of data: how it is gathered, who is compensated, and who retains influence over its use \citep{reitmaier2022opportunities}.

Looking ahead, \OmniASR can serve as a foundation for broader research agendas that connect ASR to multimodal AI, language preservation, and participatory technology governance. Future directions include combining \OmniASR with large language models to support conversational agents in under-resourced languages, embedding it in community-run archives to keep linguistic data locally controlled, and expanding its role in speech translation technologies. At the same time, sustaining open multilingual resources at this scale will require policymakers, funders, and interdisciplinary researchers to confront how to share responsibility for building and maintaining them in ways that prioritize long-term community needs \citep{wang2024human}. By situating innovation within these broader ethical and institutional contexts, \OmniASR seeks not only to advance the state-of-the-art but also to reshape the terms of engagement for how the next generation of community-focused AI will be built, shared, and governed.

\clearpage
\newpage
\bibliographystyle{assets/plainnat}
\bibliography{paper}

\clearpage
\newpage
\beginappendix
\section{\OmniASR Language Coverage}
\label{sec:appending}

{
    \tiny\renewcommand{\arraystretch}{.8}
    \begin{adjustbox}{max width=\textwidth}
\begin{tabular}{llc}
        \toprule
        \textbf{Code} & \textbf{Name} & \textbf{Res}\\
        \midrule
        \texttt{aae\_Latn} & Arbëreshë Albanian & L \\
        \texttt{aal\_Latn} & Afade & L \\
        \texttt{abb\_Latn} & Bankon & L \\
        \texttt{abi\_Latn} & Abidji & M \\
        \texttt{abk\_Cyrl} & Abkhazian & M \\
        \texttt{abn\_Latn} & Abua & L \\
        \texttt{abp\_Latn} & Abellen Ayta & M \\
\bottomrule
    \end{tabular}
    \quad
    \begin{tabular}{llc}
        \toprule
        \textbf{Code} & \textbf{Name} & \textbf{Res}\\
        \midrule
        \texttt{bcw\_Latn} & Bana & M \\
        \texttt{bcy\_Latn} & Bacama & L \\
        \texttt{bcz\_Latn} & Bainouk-Gunyaamolo & L \\
        \texttt{bda\_Latn} & Bayot & L \\
        \texttt{bde\_Latn} & Bade & L \\
        \texttt{bdg\_Latn} & Bonggi & M \\
        \texttt{bdh\_Latn} & Baka (South Sudan) & H \\
\bottomrule
    \end{tabular}
    \quad
    \begin{tabular}{llc}
        \toprule
        \textbf{Code} & \textbf{Name} & \textbf{Res}\\
        \midrule
        \texttt{bzh\_Latn} & Mapos Buang & M \\
        \texttt{bzi\_Thai} & Bisu & M \\
        \texttt{bzj\_Latn} & Belize Kriol English & M \\
        \texttt{bzw\_Latn} & Basa (Nigeria) & L \\
        \texttt{caa\_Latn} & Chortí & M \\
        \texttt{cab\_Latn} & Garifuna & M \\
        \texttt{cac\_Latn} & Chuj & H \\
\bottomrule
    \end{tabular}
    \end{adjustbox}
}

{
    \tiny\renewcommand{\arraystretch}{.8}
    \begin{adjustbox}{max width=\textwidth}
    \begin{tabular}{llc}
        \toprule
        \textbf{Code} & \textbf{Name} & \textbf{Res}\\
        \midrule
        \texttt{dip\_Latn} & Northeastern Dinka & M \\
        \texttt{div\_Thaa} & Dhivehi & L \\
        \texttt{dje\_Latn} & Zarma & M \\
        \texttt{djk\_Latn} & Eastern Maroon Creole & M \\
        \texttt{dmk\_Arab} & Domaaki & L \\
        \texttt{dml\_Arab} & Dameli & L \\
        \texttt{dnj\_Latn} & Dan & M \\
\bottomrule
    \end{tabular}
    \quad
    \begin{tabular}{llc}
        \toprule
        \textbf{Code} & \textbf{Name} & \textbf{Res}\\
        \midrule
        \texttt{gyz\_Latn} & Geji & L \\
        \texttt{had\_Latn} & Hatam & M \\
        \texttt{hag\_Latn} & Hanga & M \\
        \texttt{hah\_Latn} & Hahon & L \\
        \texttt{hak\_Latn} & Hakka Chinese & M \\
        \texttt{hao\_Latn} & Hakö & L \\
        \texttt{hap\_Latn} & Hupla & H \\
\bottomrule
    \end{tabular}
    \quad
    \begin{tabular}{llc}
        \toprule
        \textbf{Code} & \textbf{Name} & \textbf{Res}\\
        \midrule
        \texttt{kfb\_Deva} & Northwestern Kolami & M \\
        \texttt{kff\_Telu} & Koya & M \\
        \texttt{kfk\_Deva} & Kinnauri & L \\
        \texttt{kfq\_Deva} & Korku & L \\
        \texttt{kfr\_Gujr} & Kachhi & L \\
        \texttt{kfw\_Latn} & Kharam Naga & M \\
        \texttt{kfx\_Deva} & Kullu Pahari & M \\
\bottomrule
    \end{tabular}
    \end{adjustbox}
}

{
    \tiny\renewcommand{\arraystretch}{.8}
    \begin{adjustbox}{max width=\textwidth}
    \begin{tabular}{llc}
        \toprule
        \textbf{Code} & \textbf{Name} & \textbf{Res}\\
        \midrule
        \texttt{llg\_Latn} & Lole & L \\
        \texttt{lln\_Latn} & Lele (Chad) & H \\
        \texttt{lme\_Latn} & Pévé & M \\
        \texttt{lnd\_Latn} & Lundayeh & M \\
        \texttt{lns\_Latn} & Lamnso' & M \\
        \texttt{lnu\_Latn} & Longuda & L \\
        \texttt{loa\_Latn} & Loloda & L \\
\bottomrule
    \end{tabular}
    \quad
    \begin{tabular}{llc}
        \toprule
        \textbf{Code} & \textbf{Name} & \textbf{Res}\\
        \midrule
        \texttt{mqy\_Latn} & Manggarai & L \\
        \texttt{mri\_Latn} & Maori & M \\
        \texttt{mrj\_Cyrl} & Western Mari & M \\
        \texttt{mrr\_Deva} & Maria (India) & L \\
        \texttt{mrt\_Latn} & Marghi Central & L \\
        \texttt{mrw\_Latn} & Maranao & M \\
        \texttt{msh\_Latn} & Masikoro Malagasy & L \\
\bottomrule
    \end{tabular}
    \quad
    \begin{tabular}{llc}
        \toprule
        \textbf{Code} & \textbf{Name} & \textbf{Res}\\
        \midrule
        \texttt{ory\_Orya} & Odia & H \\
        \texttt{oss\_Cyrl} & Ossetian & M \\
        \texttt{ote\_Latn} & Mezquital Otomi & M \\
        \texttt{otq\_Latn} & Querétaro Otomi & M \\
        \texttt{ozm\_Latn} & Koonzime & M \\
        \texttt{pab\_Latn} & Parecís & M \\
        \texttt{pad\_Latn} & Paumarí & M \\
\bottomrule
    \end{tabular}
    \end{adjustbox}
}

{
    \tiny\renewcommand{\arraystretch}{.8}
    \begin{adjustbox}{max width=\textwidth}
    \begin{tabular}{llc}
        \toprule
        \textbf{Code} & \textbf{Name} & \textbf{Res}\\
        \midrule
        \texttt{rwr\_Deva} & Marwari (India) & L \\
        \texttt{sab\_Latn} & Buglere & M \\
        \texttt{sag\_Latn} & Sango & M \\
        \texttt{sah\_Cyrl} & Yakut & M \\
        \texttt{saj\_Latn} & Sahu & M \\
        \texttt{saq\_Latn} & Samburu & M \\
        \texttt{sas\_Latn} & Sasak & M \\
\bottomrule
    \end{tabular}
    \quad
    \begin{tabular}{llc}
        \toprule
        \textbf{Code} & \textbf{Name} & \textbf{Res}\\
        \midrule
        \texttt{tfr\_Latn} & Teribe & M \\
        \texttt{tgc\_Latn} & Tigak & L \\
        \texttt{tgj\_Latn} & Tagin & H \\
        \texttt{tgk\_Cyrl} & Tajik & H \\
        \texttt{tgl\_Latn} & Tagalog & L \\
        \texttt{tgo\_Latn} & Sudest & M \\
        \texttt{tgp\_Latn} & Tangoa & M \\
\bottomrule
    \end{tabular}
    \quad
    \begin{tabular}{llc}
        \toprule
        \textbf{Code} & \textbf{Name} & \textbf{Res}\\
        \midrule
        \texttt{vmz\_Latn} & Mazatlán Mazatec & L \\
        \texttt{vro\_Latn} & Võro & L \\
        \texttt{vun\_Latn} & Vunjo & M \\
        \texttt{vut\_Latn} & Vute & M \\
        \texttt{wal\_Ethi} & Wolaytta & M \\
        \texttt{wal\_Latn} & Wolaytta & M \\
        \texttt{wap\_Latn} & Wapishana & H \\
\bottomrule
    \end{tabular}
    \end{adjustbox}
}
\captionof{table}{Full list of languages supported by \OmniASR, including language code, English name, and resource level (Low, Medium, High).}\label{tab:full_lang_list}

\section{WER Filtering}
\label{sec:appendix_wer_filtering}

WER-thresholds were used to filter out samples likely to be of low quality from the \vendorcorpus{} ASR dataset. Values ranged from 150 to 250 WER. These were determined qualitatively and selected to filter out samples with obviously misaligned audio/text. For example:

\hfill

\begin{verbatim}
Reference:
okoro ekwup mmotima nson wo mawanne ochike machip akpan pimoruku bebogye
Hypothesis:
okoro ekwu otok kpena kpe fu bok obo mo tim so woma wane mo chike ma achit
akpe pa mo orugo be boya bep be bae bake bonga akpe pe nok boya

Reference:
en sa w konn sa k pase
Hypothesis:
en fò w konn sa k pase n ap tou benefisye yon staj men m byen kwè so kò
kòman kote sa ye lankò menm chak ki bay bon moun yo wi me nm ja ou ka

Reference:
enh se fèt dè mè se fèt ou ankò
Hypothesis:
elepicit m konnen lepichit m konnen wi m konnen demis li rele en skisoee
bon tetout fason pann fèt aa o byen pete ye e fèe fèt b  èmè pis fèt ou ankò
\end{verbatim}

\hfill

In the above examples, it is clear in listening to the audio that the hypotheses generated by our model are more accurate than the reference texts, so we filtered such examples out.

\section{Prompts and Guidelines for Commissioned Data Collection}
This section contains the recording prompts and transcription guidelines for our commissioned data collection.
\subsection{Recording guidelines}\label{appx:guideline:record}
\begin{itemize}
    \item Please record in a quiet environment.
    \item During the recording, please refrain from: 
    \begin{itemize}
        \item touching the microphone,
        \item blowing into the microphone,
        \item moving things around that are close to the recording device.
    \end{itemize}
    \item Please refrain from clearing your throat, coughing, sneezing, or making any loud sounds during the recording.
    \item Please refrain from eating or drinking during the recording.
    \item Please speak in a natural, normal voice.
    \item Please speak at a normal pace and not too quickly or too slowly.
    \item If you encounter names and words that are in a different language (for example, an English name when you are speaking Swahili), please do your best to pronounce the name as you normally would in the target language.
    \item Please refrain from sharing any personally identifiable information in the recordings, whether it pertains to you or others. Personally identifiable information includes:
    \begin{itemize}
        \item Full name
        \item Phone number
        \item Home address
        \item Email address or other account identifiers, such as social media handles
        \item Passport number
        \item Social Security Number or analogous identification numbers
        \item Health information 
        \item Sexual orientation
        \item Political affiliation
        \item Any other analogous information
    \end{itemize}
\end{itemize}

\subsection{Transcription guidelines}\label{appx:guideline:transcript}
Your job is to transcribe exactly what was said in the recording, including a representation of all the disfluencies and noises it contains.
\begin{itemize}
    \item If the recording contains grammatical mistakes, these should not be corrected in the transcription.
    \item The only characters allowed in the transcription are letters of the given language, punctuation and the set of special tags specified below.
    \item (Updated) Wherever possible and if this is applicable to your language, please use punctuation in transcripts as you would normally do in your written language. Please also capitalize the beginnings of new sentences if applicable.
\end{itemize}

\paragraph{Numbers and acronyms.}
\begin{itemize}
    \item Numbers should be spelled out in words. They should not be written in the numeral system.
    \begin{itemize}
        \item \textcolor{red}{\textbf{Incorrect}}: \textit{I walked exactly 2017 steps.}
        \item \textcolor{ForestGreen}{\textbf{Correct}}: \textit{I walked exactly two thousand seventeen steps.}
    \end{itemize}
    \item Acronyms should be written as they are normally written in the language, following standard capitalization rules. They should not be transcribed phonetically.
    \begin{itemize}
        \item \textcolor{red}{\textbf{Incorrect}}: \textit{They were arrested by the eff bee eye last Thursday.}
        \item \textcolor{ForestGreen}{\textbf{Correct}}: \textit{They were arrested by the FBI last Thursday.}
    \end{itemize}
\end{itemize}

\paragraph{Punctuation and symbols}
\begin{itemize}
    \item Use the punctuation that is appropriate for writing in the given language.
    \item Symbols for currencies, percentages, etc. should be avoided, and should instead be spelled out.
    \begin{itemize}
        \item  \textcolor{red}{\textbf{Incorrect}}: \textit{This bag cost me only \$10!}
        \item \textcolor{ForestGreen}{\textbf{Correct}}: \textit{This bag cost me only ten dollars!}
    \end{itemize}
\end{itemize}

\paragraph{Special tags}

The following special tags should be used to mark disfluencies, fillers, and other types of non-verbal content.
\begin{table}[!ht]
    \centering
    \begin{tabular}{p{3cm}p{12cm}}
    \toprule
    \textbf{Tag} & \textbf{Meaning} \\ \midrule
    \texttt{<laugh>} & The sound of laughter. \\ 
\midrule
\texttt{<hesitation>} & A hesitation sound, often used by speakers while thinking of the next thing to say. In English, some common hesitation sounds are “err”, “um”, “huh”, etc. \\ 
\midrule
\texttt{<unintelligible>} & A word or sequence of words that cannot be understood.\\
\midrule
\texttt{<noise>} & Any other type of noise, such as the speaker coughing or clearing their throat, a car honking, the sound of something hitting the microphone, a phone buzzing, etc. \\
    \bottomrule
    \end{tabular}
    \caption{Special tags used for transcription}
    \label{tab:transcript_tags}
\end{table}
\begin{itemize}
    \item Tags should be inserted in the transcription at the appropriate location, and should be separated from the other content by spaces; for example:
    \begin{itemize}
        \item \textit{And then I \texttt{<noise>} went on holiday.}
        \item \textit{Well, \texttt{<noise>} \texttt{<laugh>} it wasn’t exactly a holiday \texttt{<laugh>}}
    \end{itemize}

\item When we speak, we often insert hesitations while thinking of the next idea we want to say. Some common hesitations in English are “err”, “um” and “uh”. Since these hesitations can vary significantly in the exact sounds and length used, and often there are no clear rules on how they should be written, for this project they should all be represented using the tag \texttt{<hesitation>}. Only this tag should be used. You should not attempt to transcribe hesitations using letters, such as “err”.
\end{itemize}

\paragraph{Word segments, false starts and repeated words.} \begin{itemize}
    \item Spontaneous speech naturally contains false starts where only a fragment of a full word is produced. For these instances, please transcribe to the best of your ability the word fragment and attach a hyphen at the end of the word (-) to indicate the word is a false start. 
    \begin{itemize}
        \item \textit{His name is \underline{Jo- Jona-} Jonathan.}
    \end{itemize}
    \item Sometimes speakers will repeat a word or word fragment multiple times. This should be transcribed too.
    \begin{itemize}
        \item \textit{And then I went to \underline{the the the bed-} the bedroom.}
    \end{itemize}

\end{itemize}

\paragraph{Grammatical mistakes and colloquialisms.}
\begin{itemize}
    \item Spontaneous speech will naturally contain grammatical mistakes. These should not be corrected when transcribing. The transcription should reflect the spoken content exactly.
    \item Speakers may use colloqualisms (such as, in English, “gonna”, “cuz”, etc.) which may not be considered formally correct. These should be transcribed as they are, and not changed to their more formal equivalents.
\end{itemize}

\section{Quality Assurance (QA) Guidelines}\label{appx:guideline:qa}
In this appendix, we detail the guidance provided to perform quality assurance (QA).
\subsection{Speech recording error taxonomy}
\Cref{tab:speech_qa_errors} shows the definitions used for each of the error categories. More broadly, QA technicians were asked to pay particular attention to the following speech recording issues:
\begin{itemize}
    \item General audio quality issues (e.g., volume is too low, speech is inaudible, there is constant background noise or heavy static, files seem systematically cut off before the end)
\item Ad hoc noises (e.g., rooster crowing, mechanical noises, bells or phones ringing, very long silences or pauses)
\item Other human voices (e.g., people talking in the background in the same language, or more problematic, in a different language)
\item The speaker responds to the prompts in a pivot language, not in the expected language (prompts were translated into a number of high-resource pivot languages and it can happen that the speaker will respond to the prompts in the same language as the prompts instead of responding in their native language)

\end{itemize}

\begin{table}[!ht]
    \centering
    \begin{tabular}{p{3cm}p{6cm}p{6cm}}
    \toprule
     \textbf{Category}  & \textbf{Critical example} & \textbf{Minor example} \\
     \midrule
     \small{Human vocal noise} & \small{Second voice in the background} & \small{N/A (This error is always critical)} \\
     & \small{Singing in the background} &  \\
    \midrule
    \small{Cutoff} & \small{Speech is cut off at either end of the recording} &  \small{N/A (This error is always critical)}\\
     \midrule
     \small{Background noise} & \small{Rooster crowing} & \small{Occasional mild coughing} \\
     & \small{Street noise, car honking} & \small{Occasional mild coughing} \\
     & \small{Bird chirping} & \small{Mild breathing sound}\\
     & \small{Strong wind} & \\
     \midrule
\small{Audio Glitches} & \small{Serious glitches that break up speech} & \small{Mild glitch happens in between speech} \\
\midrule
\small{Static noise} & \small{Strong static noise that affects intelligibility} & \small{Mild static noise that does not affect speech} \\
\midrule
\small{Low volume} & \small{Cannot hear the speech clearly in the max volume setting} & \small{Lower than normal but still audible at max volume} \\
\midrule
\small{Inconsistent volume} & \small{Volume changing drastically} & \small{Occasional soft voice} \\
\midrule
\small{Muffled voice} & \small{Muffled voice sounds like talking behind a curtain} & \small{Audio is not crisp but does not affect intelligibility} \\
\midrule
\small{Echo} & \small{Strong echo like speaking in a cave or tunnel such that it compromises the intelligibility of words} & \small{Mild echo in non-studio environment} \\
\midrule
\small{Microphone Noise} & \small{Any hissing, plosive, popping noise that breaks the speech} & \small{Mild pop noise when turning on/off the recorder} \\
\midrule
\small{Pause / Silence} & \small{Long pauses} & \small{Short pauses when speaker is thinking} \\
 & \small{- If at the start or end of speech and above 2s} & \\
 & \small{- If at the middle of speech and above 5s} & \\
 & \small{- If more than $\frac{1}{3}$ of the audio is made up of leading/trailing silence or intra-sentential silence (excluding normal pauses between words)} & \\
\midrule
\small{Unnatural speech} & \small{Consistent stutter or mumbling} & \small{Occasional repeated words and syllable} \\
 & \small{Extremely not fluent, words uttered individually} & \\
 & \small{Whisper} & \\
 & \small{Feels like someone reading / monotonous speech} & \\
     \bottomrule
    \end{tabular}
    \caption{Description of all error categories used for speech recording in-depth quality assurance.}
    \label{tab:speech_qa_errors}
\end{table}

\subsection{Transcript error taxonomy}

\Cref{tab:transcript_qa_errors} shows the definitions used for each of the error categories. More broadly, QA technicians were asked to pay particular attention to the following transcript issues:
\begin{itemize}
\item General transcription issues (e.g., the transcript does not match the audio file at all, the transcript is in an unexpected writing system, the transcript is in the International Phonetic Alphabet, the transcript is missing words, the transcript is much shorter or longer than it should be)
\item Transcription issues that are specific to a language (e.g., a few non-Unicode-compliant characters have been used)
\item Issues related to the use of event-marking tags (a specific tag set has been defined by the project team; \Cref{tab:transcript_tags})
\end{itemize}
\begin{table}[!ht]
    \centering
    \begin{tabular}{p{3.5cm}p{6cm}p{5.5cm}}
    \toprule
     \textbf{Category}  & \textbf{Critical example} & \textbf{Minor example} \\
     \midrule
    \small{Mismatch} & \small{Transcript file does not match the audio at all (either in content or in length) } & \small{N/A (mismatch is critical)} \\
\midrule
\small{Wrong writing system} & \small{The transcript does not use the expected writing system}  & \small{N/A (writing system is critical)} \\
 & \small{The transcript is in IPA or other phonetically-based system} & \\
 & \small{Different writing standard, inconsistency in the spelling (the same word spelled in different ways)} & \\
\midrule
\small{Wrong tags} & \small{The transcript includes made-up tags} & \small{N/A (all mistaggings are critical)} \\
 & \small{Tags are not used adequately (e.g., \texttt{<noise>} instead of \texttt{<hesitation>})} & \\
\midrule
\small{Numbers} & \small{The presence of numbers written in digits} & \small{(N/A writing digits is critical)} \\
\midrule
\small{Incomplete} &  \small{The transcript is abridged rather than verbatim} & \small{The transcript seems to sometimes be missing a word or two} \\
 & \small{The transcript consistently misses words} \\
\midrule
\small{Inconsistent tagging} & \small{The tag set being used is compliant but the transcriber consistently switches between tags for the same audio events} & \small{A few tags show inconsistency, especially for borderline audio events} \\

     \bottomrule
    \end{tabular}
    \caption{Description of all error categories used for transcript in-depth quality assurance.}
    \label{tab:transcript_qa_errors}
\end{table}

\end{document}